\documentclass[lettersize,journal]{IEEEtran}
\usepackage{amsfonts,amssymb}
\usepackage{graphicx} 
\usepackage{booktabs}
\usepackage{multirow}
\usepackage[table,xcdraw]{xcolor}
\usepackage[normalem]{ulem}
\useunder{\uline}{\ul}{}

\usepackage{amsmath,amsfonts}
\usepackage{algorithmic}
\usepackage{algorithm}
\usepackage{array}
\usepackage[caption=false,font=normalsize,labelfont=sf,textfont=sf]{subfig}
\usepackage{textcomp}
\usepackage{stfloats}
\usepackage{url}
\usepackage{verbatim}
\usepackage{graphicx}
\usepackage{cite}
\begin{document}

\title{Differentially Private Pre-Trained Model Fusion using Decentralized Federated Graph Matching}

\author{Qian~Chen,~
        Yiqiang~Chen,~\IEEEmembership{Senior~Member,~IEEE,}
        Xinlong~Jiang,~\IEEEmembership{Member,~IEEE,}\\
        Teng~Zhang,~ 
        Weiwei Dai,~
        Wuliang~Huang,~
        Zhen~Yan,~ and
        Bo Ye

\thanks{Manuscript received XX XX, 2023; revised XX XX, 2023. This work was supported by Beijing Municipal Science \& Technology Commission (No.Z221100002722009), National Natural Science Foundation of China (No.62202455), Youth Innovation Promotion Association CAS, and the Science Research Foundation of the Joint Laboratory Project on Digital Ophthalmology and Vision Science (No. SZYK202201). }
\thanks{Qian Chen, Yiqiang Chen, Xinlong Jiang, Teng Zhang, Wuliang Huang, and Zhen Yan are with the Institute of Computing Technology, Chinese Academy of Sciences, Beijing, China, and also with the University of Chinese Academy of Sciences, Beijing, China. (e-mail: \{chenqian20b, yqchen, jiangxinlong, zhangteng19s, huangwuliang19b, yanzhen21s\}@ict.ac.cn).}
\thanks{Weiwei Dai is with Changsha Aier Eye Hospital, Changsha, Hunan, China (e-mail: daiweiwei@aierchina.com).}
\thanks{Bo Ye is with Nanchang Aier Eye Hospital, Nanchang, Jiangxi, China (e-mail: yebo@aierchina.com).}
\thanks{\textit{Corresponding authors: Yiqiang Chen and Xinlong Jiang.}}
}



\maketitle

\begin{abstract}
Model fusion is becoming a crucial component in the context of model-as-a-service scenarios, enabling the delivery of high-quality model services to local users. However, this approach introduces privacy risks and imposes certain limitations on its applications. Ensuring secure model exchange and knowledge fusion among users becomes a significant challenge in this setting. To tackle this issue, we propose PrivFusion, a novel architecture that preserves privacy while facilitating model fusion under the constraints of local differential privacy. PrivFusion leverages a graph-based structure, enabling the fusion of models from multiple parties without necessitating retraining. By employing randomized mechanisms, PrivFusion ensures privacy guarantees throughout the fusion process. To enhance model privacy, our approach incorporates a hybrid local differentially private mechanism and decentralized federated graph matching, effectively protecting both activation values and weights. Additionally, we introduce a perturbation filter adapter to alleviate the impact of randomized noise, thereby preserving the utility of the fused model. Through extensive experiments conducted on diverse image datasets and real-world healthcare applications, we provide empirical evidence showcasing the effectiveness of PrivFusion in maintaining model performance while preserving privacy. Our contributions offer valuable insights and practical solutions for secure and collaborative data analysis within the domain of privacy-preserving model fusion.
\end{abstract}


\section{Introduction}
\IEEEPARstart{D}{ata}-driven artificial intelligent systems have gradually evolved into a new paradigm that is model-driven \cite{foundationM}. So far, as a provider of Model-as-a-Service (MaaS), there are nearly 81k models on HuggingFace \cite{HuggingFace} and about 800 models on ModelScope \cite{ModelScope} that can be downloaded and executed in some common ways(e.g., fine-tuning \cite{tuning}, domain adaptation \cite{DA}, knowledge distillation \cite{KD}, model fusion \cite{OTfusion}, etc.) for various downstream tasks, such as finance, transportation, smart buildings, especially collaborative healthcare. Collaborative healthcare applications \cite{HealthcareSurvey} leverage artificial intelligence and medical expert knowledge to enable model deployment and adaptation across multiple healthcare organizations. Local fine-tuning of pre-trained models enables local control over medical data and allows model providers to offer assistance, as illustrated in Fig. 1. However, collaborative healthcare applications within the context of MaaS present inherent privacy risks and application limitations. Firstly, users are constrained to rely on a presumed trusted third-party entity as mandated by the MaaS framework for the fusion and sharing of model knowledge, which poses challenges in terms of practical guarantees. Additionally, uploading locally adapted models into a less trustworthy model pool can compromise the confidentiality of sensitive local medical information. Consequently, if direct data exchange is unfeasible, the pressing question arises: \textbf{Is there a decentralized approach that can ensure secure model sharing and knowledge fusion among users?}

\begin{figure}[t]
\includegraphics[width=\linewidth]{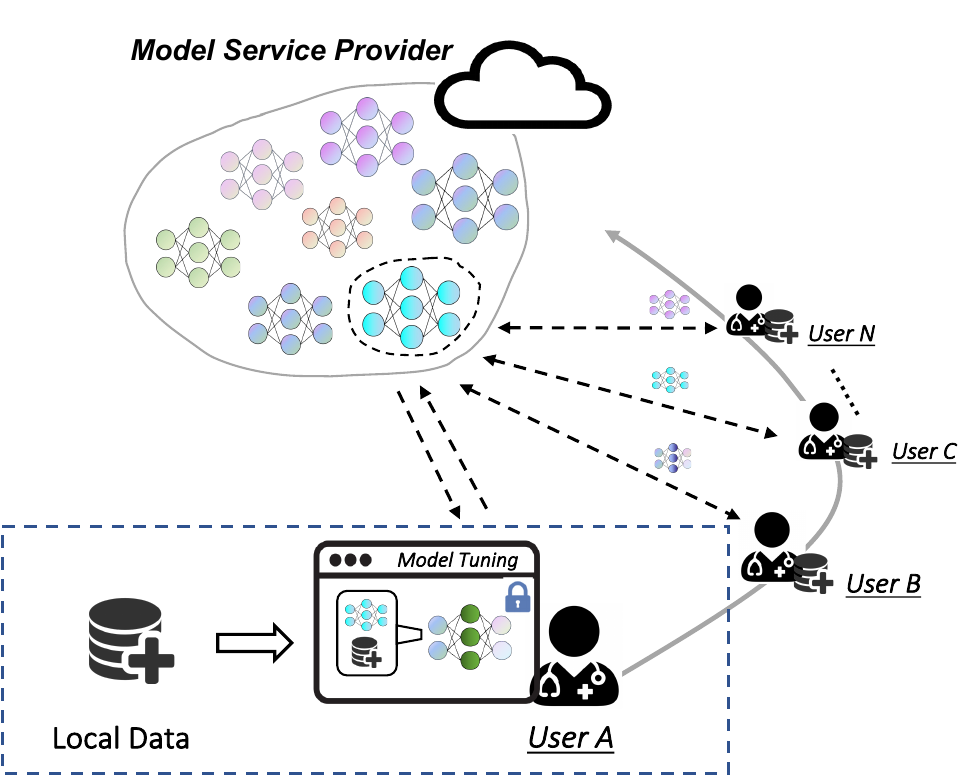}
\caption{Users select a pre-trained model from the ``Model Service Provider" and subsequently fine-tune and update it using local data. The updated model is then contributed back to the pool, enabling other users to leverage it within their own processes.}
\label{fig:res}
\end{figure}

Pre-trained model fusion emerges as a promising and feasible approach to address the aforementioned challenges. The pre-trained model fusion aims to achieve local model updates by averaging parameters with minimal training data. The redundancy of neural network parameterization makes it challenging to find a one-to-one correspondence between the weights of different networks, hindering the effectiveness of vanilla averaging, which performs poorly on trained networks with nonlinear differences in weights. To achieve better performance of the fused model, it is necessary to obtain a mapping relationship between the two neural networks by sharing their structure and parameters. This information is shared to calculate the similarity or affinity of weights to get the permutation matrix and align the neurons for layer-by-layer fusion with other models. Recently, there have been several works that have delved into the field of model fusion. Singh et al. \cite{OTfusion} suggest optimal transport to merge models from multiple sources, demonstrating its effectiveness in image classification tasks. Liu et al. \cite{GAMF} propose a deep neural network fusion approach based on graph matching to improve model ensemble and federated learning. It achieves state-of-the-art results on several datasets. Ainsworth et al. \cite{Git} present a method for merging models that are identical up to permutation symmetries, which is shown to improve performance in certain tasks. Akash et al. \cite{akash2022wasserstein} use Wasserstein Barycenter to fuse multiple models and explore the linear mode connectivity of neural networks in the context of model fusion. Jin et al. \cite{jin2022dataless} present a dataless knowledge fusion approach for language models, which combines the weights of multiple pre-trained models to improve their performance on downstream tasks. In scenarios involving cross-institutional collaboration, these methods inevitably require sharing information about the model, including its neural network structure, weights, and hyper-parameters, to varying degrees. Compared to traditional ensemble-based methods \cite{bagging,stacked,boosting}, pre-trained model fusion can avoid the need to save all trained models, thereby reducing the space resources required. Compared to distillation-based methods \cite{KD, DFKD}, by incorporating a teacher network as a guide, the process of pre-trained model fusion can be accomplished without the need for retraining using training samples or pseudo samples. Compared to federated learning methods \cite{FedMA, FL2}, pre-trained model fusion can effectively address the dilemma of unavailable model training samples. Considering the current technological advantages, the integrated and updated model has the potential to be shared among multiple healthcare organizations, fostering a positive cycle of model development and improvement within the MaaS community.

However, privacy poses a substantial challenge in the domain of real-world applications and decentralized model sharing, particularly when it comes to the exchange of models among their respective owners. Even during the process of model fusion, sharing the structure and weights of neural networks can still lead to a privacy breach of local data. This can occur through various types of attacks, including model extraction attacks \cite{extraction1,extraction2}, model inversion attacks \cite{inversion1,inversion2}, or membership inference attacks \cite{membership1,membership2}. The sensitivity of healthcare data and legal constraints on data sharing highlights the necessity for healthcare models to incorporate privacy protection measures. To address the privacy challenges arising from knowledge sharing among different parties, various privacy-preserving techniques have been proposed, including Differential Privacy (DP) \cite{DP, DP_06}, Homomorphic Encryption (HE) \cite{HE1,HE2}, Secure Multi-Party Computation (SMPC) \cite{SMPC1,SMPC2}, and hybrid methods \cite{hybrid1,hybrid2}. DP provides a strong privacy guarantee by adding noise to the output of a computation, making it difficult for attackers to extract sensitive information from the result. HE allows computation on encrypted data without decrypting it, which can prevent unauthorized access to sensitive data. SMPC enables multiple parties to jointly compute a function while keeping their inputs private, but it can be computationally expensive and communication intensive. Hybrid methods combine the benefits of multiple privacy-preserving techniques and can achieve better performance than individual methods alone. For privacy-preserving model fusion, DP and its variants are suitable choices as it provides a rigorous privacy guarantee with a relatively low computational cost compared to other methods. DP is well-suited for scenarios where healthcare data is sensitive and model privacy protection is critical. In addition, DP has been widely adopted and extensively studied in the privacy community, making it a mature and trusted technique for privacy preservation.

In order to address the aforementioned issues and challenges, we propose a novel decentralized architecture, PrivFusion, for model fusion that prioritizes privacy preservation. The PrivFusion addresses the challenge of balancing privacy and utility in sharing trained models. To achieve this, we adopt a graph-based representation of the trained neural network, where nodes represent neurons/channels and edges represent the weights connecting them. By leveraging a decentralized federated approach and employing randomized graph matching mechanisms that adhere to the principles of local differential privacy, models from different parties can be effectively fused into a unified model with a consistent structure. This approach establishes trust among multiple parties involved in the sharing process and promotes secure collaboration. The main contributions of this paper are summarized as follows:

\begin{itemize}
\item[$\bullet$] We propose PrivFusion, a novel and privacy-preserving architecture for decentralized model fusion across multiple parties. To the best of our knowledge, we are the first to address the privacy concerns associated with the fusion of pre-trained models.
\item[$\bullet$] We propose a decentralized federated graph matching approach that aligns neurons of neural networks layer by layer to ensure effective model fusion. Moreover, we achieve a trade-off between privacy and utility by leveraging a hybrid local differentially private perturbation mechanism.
\item[$\bullet$] We integrate model fusion strategies based on activation values and weights, analyzing their contributions to the fused model through individual perturbed performance assessment. To improve the utility of private model fusion, we employ a perturbation filter adapter to alleviate the impact of randomized noise on the fused model's performance. 
\item[$\bullet$] Experiments show the effectiveness of PrivFusion in both privacy goals and model performance. Additionally, we have also validated the PrivFusion in some real-world collaborative health tasks. 
\end{itemize}

The rest of this paper is organized as follows. Section 2 provides a brief review of related work. Section 3 introduces preliminaries and problem formulation, while Section 4 presents the detail of the PrivFusion. Section 5 provides a detailed explanation and analysis of the experimental results. Finally, Section 6 concludes the paper.

\section{Related Work}
\subsection{Model Fusion} 
Model fusion or ensemble methods have garnered significant attention due to their capacity to enhance the performance of machine learning models. Traditional ensemble learning is a well-established approach that combines multiple models to achieve superior performance compared to individual models. This technique entails training a collection of base models on the same dataset using diverse algorithms or settings and aggregating their predictions to generate a final prediction. Prominent ensemble methods include bagging \cite{bagging}, stacking \cite{stacked}, and boosting \cite{boosting}, each employing distinct techniques for combining model predictions. Bagging involves training multiple models on different subsets of the data and combining their predictions through averaging or voting. Boosting, on the other hand, trains models sequentially, with each model emphasizing examples that previous models struggled with. Stacking, a versatile approach, combines predictions from multiple models using another model, known as a meta-learner, which learns to weigh the predictions based on their accuracy. These ensemble methods offer various strategies for aggregating model predictions and can be applied to a wide range of tasks in different domains.

In recent years, several methods have been proposed to address scenarios where training data is unavailable due to privacy concerns or other reasons. One notable approach is OTFusion \cite{OTfusion}, which tackles model fusion as a linear assignment problem focusing on weight alignment. By utilizing Wasserstein barycenters, OTFusion improves upon simple averaging methods. However, it has a limitation in neglecting second-order weight similarity, leading to potential degradation in performance. To address this limitation, Liu et al. \cite{GAMF} propose GAMF, a model fusion method that formulates the problem as graph matching. GAMF aligns channels in each layer to maximize weight similarity, taking into account second-order similarities. This efficient graduated assignment-based method overcomes challenges associated with the quadratic assignment problem. GAMF has been extended to handle multi-model fusion and has demonstrated superior performance compared to state-of-the-art baselines in tasks such as compact model ensemble and federated learning across various datasets. In parallel, Ainsworth et al. \cite{Git} introduce three algorithms that permute hidden units of one model to align with a reference model, effectively merging them in weight space. This process generates functionally equivalent weights within an approximately convex basin. Subsequent works, such as \cite{akash2022wasserstein} and \cite{jin2022dataless}, have built upon these foundations. Akash et al. propose a fusion method that employs Wasserstein distance to combine multiple models, while Jin et al. focus on merging weights of language models to enhance their capabilities. These advancements in model fusion techniques have contributed to enhancing the performance of pre-trained models, addressing challenges related to unavailable training data.

In summary, the selection of a model fusion or ensemble approach should consider the specific requirements of the problem, including privacy considerations, data availability, and computational resources. It is essential to prioritize the safeguarding of sensitive information related to private models, such as neural network architectures, weights and proprietary data, throughout the model fusion process.
\subsection{Local Differential Privacy} 
Differential Privacy \cite{DP,DP_06} is a robust framework for quantifying the privacy guarantees offered by data release mechanisms. It provides a rigorous and provable definition of privacy by ensuring that an individual's data remains indistinguishable, regardless of their participation in a dataset. This framework has gained significant attention as a rigorous and well-founded privacy definition, leading to extensive research in recent years. Local Differential Privacy (LDP) \cite{LDP,DP_06} is a specific instantiation of differential privacy that has emerged as a popular method for preserving privacy while collecting sensitive data and computing statistical queries. LDP enables the calculation of aggregate functions, such as mean, count, and histogram, while preserving the privacy of individual data points. Notably, major technology companies including Google, Apple, and Microsoft have embraced LDP for conducting large-scale private data analytics \cite{LDP_survety}. The adoption of differential privacy and LDP has paved the way for conducting privacy-preserving data analysis, ensuring the protection of individuals' sensitive information while enabling valuable insights to be derived from the data. The rigorous nature of these privacy-preserving techniques offers a strong foundation for addressing privacy concerns in data-driven applications and has motivated further advancements in the field.

LDP operates by introducing random noise to the data prior to its disclosure, rendering it challenging for an attacker to deduce the original data. The application of LDP has been extensively explored in the context of graph-structured data representations, particularly when pre-trained models can be transformed into such representations. Protecting sensitive information in graphs has garnered increasing attention in recent years, leading to the exploration of differential privacy (DP) and LDP in this domain. One common strategy involves utilizing graph-based mechanisms to inject noise into various aspects of the data, including node features, graph structure, and edge weights \cite{AsgLDP,2013LDPgraph,2020LDPgraph}. This approach aims to preserve the privacy of graph data while allowing for meaningful analysis. Another approach leverages Graph Neural Networks (GNNs) to learn representations of the data, enabling downstream tasks without compromising sensitive information \cite{fedgnn, LDPGNN, Graph_LDP}. In this approach, the randomization of the graph occurs without explicitly focusing on the model optimization process. These advancements in applying DP and LDP to graph data have opened up new avenues for privacy-preserving analysis, enabling the protection of sensitive information while extracting valuable insights from graph-structured datasets. The integration of LDP and GNNs showcases the potential for achieving a balance between privacy preservation and effective data utilization in graph-related applications.

\begin{figure*}[t]
\includegraphics[width=\textwidth, scale=1.00]{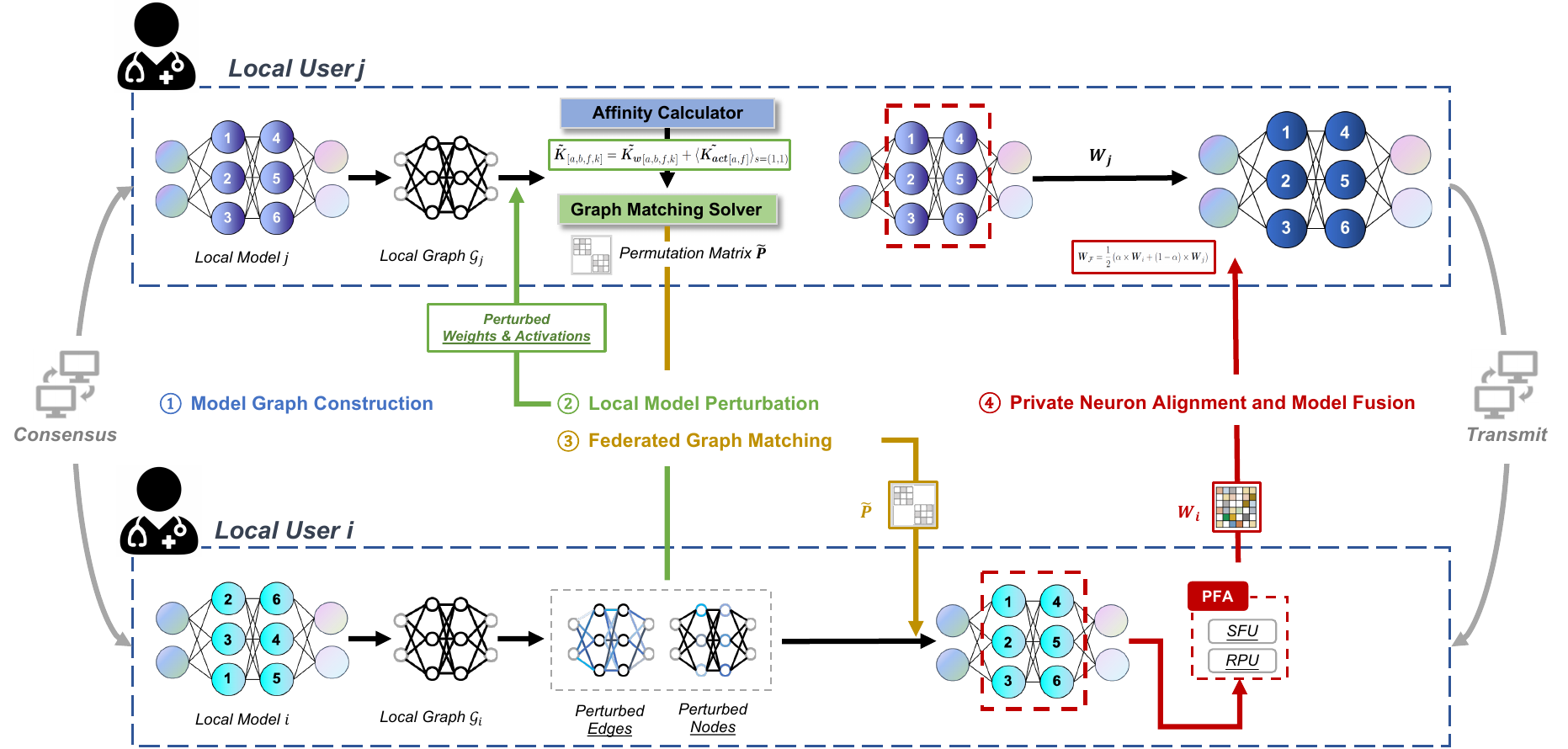}
\caption{The framework and workflow of the PrivFusion.}
\label{fig:res2}
\end{figure*}

\section{Preliminaries}
\subsection{Notations and Problem Formulation}
We formulate the pre-trained model fusion as a graph matching and publishing problem without losing generality. Many neurons in a trained neural network model can interact with one another and recognize a certain pattern. Based on this association structure, the model may be viewed as data with a graph structure, naturally, sharing models are turned into transmitted graph data.

The trained neural network can be described by a graph in which nodes represent the neurons, edges represent the relations between them, and features associated with the edges represent the weights of the neural network. We firstly construct $ m $ trained neural networks as graphs $ \mathcal{G}_{1}, \mathcal{G}_{2}, \ldots, \mathcal{G}_{m} $. Each graph $\mathcal{G}_{i}$ contains $ n_{i} $ nodes, and its neighboring layers are fully connected. For a given graph $ \mathcal{G}_{i} = (\mathcal{\bf V}_{i}, \mathcal{\bf X}_{i}, \mathcal{\bf A}_{i}) $, $\mathcal{\bf V}_{i} $ is the node/neuron set, and $ \mathcal{\bf X}_{i} \in \mathbb{R}^{N\times d}$ denotes $d$-dimensional node features of the ${i}^{th}$ network. The node features can be considered for activations of neurons by running inference over a set of samples from different classes. $ \mathcal{\bf A}_{i} \in \mathbb{R}^{N \times N} $ represents the adjacency matrix encodes the weighted connectivity of the graph $\mathcal{G}_{i}$. The adjacency matrix is relatively sparse since the cross-layer links are meaningless except for the connections of neurons in the neighboring layer. Obviously, $\mathcal{G}_{i}$ is a directed graph whose direction is pointed at the model from input to output. 

Training the neural network with stochastic gradient descent and different training sets shuffles the neurons. Furthermore, the model fused by simply averaging is ineffective if the weights of neurons are in unaligned positions. Since graph matching can solve the alignment problem of neuron positions, for simplicity, we formulate the matching and fusing with two fully connected networks with two hidden layers without bias. The goal of matching is to find a suitable permutation of the nodes(neurons or channels) between any two graphs $\mathcal{G}_{i}, \mathcal{G}_{j}$, $i, j \in 1,2, \ldots, m$, which is equivalent to permuting $\mathcal{G}_{i}$ to fit $\mathcal{G}_{j}$ by graph matching. For example, fusing two fully-connected networks through graph matching can be represented as follows:
\begin{equation}
\max _{\boldsymbol{P}} \sum_{a=0}^{d_{\Sigma}-1} \sum_{b=0}^{d_{\Sigma}-1} \sum_{e=0}^{d_{\Sigma}-1} \sum_{f=0}^{d_{\Sigma}-1} \boldsymbol{P}_{[a, b]} \boldsymbol{K}_{[a, b, f, k]} \boldsymbol{P}_{[f, k]} 
\end{equation}
where $\boldsymbol{P}$ is the permutations of two hidden layers, and the neurons of input/output layers are assumed to have been aligned and need not be permuted. These constraints guarantee the existence of one-to-one mapping relationships between neurons that are within the same layer. The similarity of two graphs can be denoted as $\boldsymbol{K} \in  \mathbb{R}^{{d_{\Sigma}} \times {d_{\Sigma}} \times {d_{\Sigma}} \times {d_{\Sigma}}}$, which is a 4-dimensional affinity tensor. Element $\boldsymbol{K}_{[a, b, f, k]}$ measures the affinity between the edges $(a, f)$ and $(b, k)$ and denotes the similarity between the weight matrices of two graphs in the model fusion problem. The measurement of similarity can be done by common means such as Gaussian kernel, cosine similarity, correlation distance, etc.

\subsection{Privacy Threat and Protection}
In a decentralized architecture, where there is no involvement of a trusted third-party server, model owners (local users) are considered semi-trusted, meaning they are honest but curious. The adversary attempts to extract model details and speculate on private training samples. Consequently, it is unsafe to directly share the actual weight values for calculating affinity. Our ultimate goal is to obtain a set of mechanisms that allows model owners to aggregate multiple models' knowledge from other parties while preserving the privacy of the models. Additionally, model owners should be able to benefit from other participants over the collected noisy graph and improve their own models with the best possible generalization capability after optimizing. To ensure the privacy of the entire process of model fusion, we use a privacy-preserving scheme based on differential privacy, which is defined below.

As a rigorous and provable definition of privacy, differential privacy \cite{DP} has received a lot of attention and has been extensively researched in the last few years. Differential Privacy provides privacy protection for data with adversary background knowledge maximization, and it is a notion of privacy that addresses the privacy problem of statistical databases.

{\bf\textit{Definition 1 ($\epsilon$-Differential Privacy)}}: \textit{A random mechanism $\mathcal{M}$ satisfies $\epsilon$-differential privacy if for any pair of neighboring datasets $X$ and $X'\in X$ that differ by a single data instance and for any set of outcomes $S \subseteq R$,}
\begin{equation}
\boldsymbol{Pr}[\mathcal{M}(X) \in \mathcal{S}] \leq \exp (\epsilon) \boldsymbol{Pr}\left[\mathcal{M}\left(X^{\prime}\right) \in \mathcal{S}\right]
\end{equation}
where $\epsilon$ is referred to as the privacy budget, and it quantifies the privacy risk of mechanism $\mathcal{M}$. The parameter $\epsilon$ is also used to tune utility versus privacy: a smaller (resp. larger) $\epsilon$ leads to stronger (weaker) privacy guarantees, but lower (higher) utility. $\boldsymbol{Pr}[\cdot]$ represents the randomness of the $\mathcal{M}$ on the datasets $X$ and $X'$. This definition ensures that the presence or absence of an individual will not significantly affect the output. Differential privacy has some lemmas. In the differentially private computing \cite{DP}, we can take advantage of the following lemmas: 

{\bf\textit{Lemma 1 Post-processing}}. \textit{Any calculation of output under differential privacy does not increase privacy loss.}

{\bf\textit{Lemma 2 Serialized combination theorem}}. \textit{Serialized combination of differential privacy mechanisms still satisfies differential privacy protection.}

One major problem with the notion of differential privacy is that users still have to trust a central authority to keep their privacy. This is not feasible in the setting of the cross-client model fusion problem. In order to be able to give users stronger privacy guarantees the concept of local differential privacy(LDP) was introduced. We use the definition of LDP given by \cite{LDP}. 

{\bf\textit{Definition 2 ($\epsilon$-Local Differential Privacy)}}: \textit{A random algorithm $\mathcal{A}$ satisfies $\epsilon$-local differential privacy if for all pairs of client’s values $V_{1}$ and $V_{2}$ and for all $Q\subseteq$ Range$(\mathcal{A})$ and for $\epsilon > 0$, the follow equation holds:}

\begin{equation}
\boldsymbol{Pr}\left[\mathcal{A}\left(V_{1}\right) \in Q\right] \leq \exp (\varepsilon) \boldsymbol{Pr}\left[\mathcal{A}\left(V_{2}\right) \in Q\right]
\end{equation}
Range$(\mathcal{A})$ is the set of all possible outputs of the randomized algorithm $\mathcal{A}$. 

{\bf\textit{Properties}}. Given the constraint of local differential privacy (LDP), there exists a lack of trust among users. To address this, a perturbation mechanism is applied by each user to their data before sharing it with the curator or other users. This mechanism ensures differential privacy, preserving the privacy of each individual's data. While the shared data may not have any meaningful interpretation on its own, it can be used for aggregating parameters and fusing models. Importantly, the privacy guarantee remains intact even if there is post-processing of the algorithm's output, as long as it adheres to the principles of LDP.

\section{PrivFusion: our proposed method}
This section describes the main components of our proposed method, called PrivFusion, toward differentially private model fusion via decentralized federated graph matching with a decentralized architecture. The overall workflow of the proposed PrivFusion(shown in Fig 2). Once users reach a consensus on model tasks, structure, and requirements, the process mainly consists of three components: (1) \textit{Local Model Perturbation}, (2) \textit{Decentralized Federated Graph Matching}, and (3) \textit{Private Neuron Alignment and Fusion}. The first component aims to perform privacy-preserving obfuscation on the graph that is converted by the pre-trained model. Each user(model owner) independently shares a variant of the model - the obfuscated graph data, including the obfuscated version of node features and the obfuscated weight matrix. The second component involves calculating layer-wise affinity on the noisy graph composed of obfuscated data shared by all users. It employs federated interactions to perform graph matching and obtain permutation matrices for neurons/channels. The third component involves applying neuron permutation based on the permutation matrix to align the neurons of each layer in the models. This is followed by averaging the corresponding weights of the models under LDP constraints. At the end of the entire workflow, the fused model is transmitted among the users.

In what follows, we first introduce PrivFusion's technical details, including the hybrid randomized mechanisms used for preserving model privacy that satisfies local differential privacy, as well as the decentralized federated graph matching approach. Then, we theoretically and empirically analyze the privacy-utility trade-off issue caused by the hybrid differentially private mechanisms.

\subsection{Local Model Perturbation}
In the setting of the PrivFusion, model owner ${U}_{i}$ and ${U}_{j}$ hold their local model and construct them into two graphs $\mathcal{G}_{i}$ and $\mathcal{G}_{j}$ with nodes and weighted unidirectional edges. Take the ${U}_{i}$ as an example, graph $\mathcal{G}_{i}$ includes node feature $\mathcal{\bf X}_{i}$, the adjacency matrix $\mathcal{\bf A}_{i}$ and weight matrix $\mathcal{\bf W}_{i}$. All nodes are connected layer by layer, they are only connected to nodes of neighboring layers, and there are no second-order neighbors of nodes(cross-layer connection). The mechanisms utilized to ensure the privacy of the node feature, the adjacency matrix, and the weight matrix of the model, respectively, are described in the paragraphs that follow.

{\bf Node Feature Perturbation}. We retain the activations for all of the model's neurons after running inference over a few instances. After that, we consider the corresponding activation values as features of the nodes and randomize these features. In this phase, we apply a variant of the Laplace mechanism\cite{Laplace_LDP}, one of the implementation mechanisms for LDP, to secure the delivery process of the model's activations. The privacy-preserving module adds Laplacian noise to the node feature $\boldsymbol{X}_{i}$ which is formulated as follows:

\begin{equation}
\boldsymbol{\tilde X}_{i} = \boldsymbol{X}_{i} + \textit { Laplace }\left({S(f)} / {\epsilon_{x}}\right)
\end{equation}
where $\epsilon_{x}$ is the ``privacy budget" for node feature perturbation. $S(f)$ represents $L_{1}$ sensitivity which is defined as follows.

{\bf\textit{Definition 3 ($L_{1}$ Sensitivity)}}: \textit{The $L_{1}$ sensitivity of a function $f: D^{n} \rightarrow \mathbb{R}^{d}$ is the smallest number $S(f)$ such that for all $\boldsymbol{X}_{i}, \boldsymbol{X'}_{i} \in D^{n}$ which differ in a single entry, }

\begin{equation}
\left\|f(\boldsymbol{X}_{i})-f\left(\boldsymbol{X}_{i}^{\prime}\right)\right\|_{1} \leq S(f)
\end{equation}
where $\left\| \cdot \right\|_{1}$ represents $L_{1}$ norm of a vector. Sensitivity is the key parameter that determines the amount of noise added, which refers to the maximum change caused by deleting any record. Moreover, \textit {Laplace} $(\cdot)$ is a random variable that has the following probability density function:

\begin{equation}
\boldsymbol{Pr}[x, \lambda]=\frac{1}{2 \lambda} \cdot e^{-\frac{|x|}{\lambda},} \quad \forall x \in \mathbb{R}^{d}
\end{equation}
where $x$ presents the specific variable $\lambda$ control the strength of protection (volume of noise) with a privacy budget $\epsilon_{x}$ bounded by $\frac{S(f)}{\lambda}$.

{\bf\textit{Theorem 1.}} \textit{For an arbitrary adversary, let $f(\boldsymbol{X}_{i}) : D^{n} \rightarrow \mathbb{R}^{d}$ be its query function as a means of observation. If $\lambda = maxS(f)/\epsilon_{x}$, the the mechanism $\mathcal{M}_{x}$ satisfies $\epsilon_{x}$-local differential privacy.}

{\bf\textit{Proof.}} Considering that $\boldsymbol{X}_{i}$ and $\boldsymbol{X'}_{i}$ only differ in single entry. Using the law of conditional probability and given any output $Sx = (Sx_{1}, Sx_{2},\ldots, Sx_{n})$ from the node feature perturbation mechanism $\mathcal{M}_{x}$, 

\begin{equation}
\frac{\boldsymbol{Pr}\left[\mathcal{M}_{x}(\boldsymbol{X}_{i})=Sx\right]}{\boldsymbol{Pr}\left[\mathcal{M}_{x}\left(\boldsymbol{X}_{i}^{\prime}\right)=Sx\right]}=\prod_{h} \frac{\boldsymbol{Pr}\left[\mathcal{M}_{x}(\boldsymbol{X}_{i, h})=Sx_{h} \right]}{\boldsymbol{Pr}\left[\mathcal{M}_{x}\left(\boldsymbol{X}_{i, h}^{\prime}\right)=Sx_{h} \right]}
\end{equation}

For each term in the product, $h = 1, 2, \ldots, n$. The conditional distributions are therefore standard Laplacians, which means we can bind each term and its product as

\begin{equation}
\begin{aligned}
\prod_{h} \frac{\boldsymbol{Pr}\left[\mathcal{M}_{x}(\boldsymbol{X}_{i, h})=Sx_{h} \right]}{\boldsymbol{Pr}\left[\mathcal{M}_{x}\left(\boldsymbol{X}_{i, h}^{\prime}\right)=Sx_{h} \right]} &\leq \prod_{h} exp(|f(\boldsymbol{X}_{i, h}) \\
&-f(\boldsymbol{X'}_{i, h}|/\lambda) \\
& = exp(\left\|f(\boldsymbol{X}_{i})-f\left(\boldsymbol{X}_{i}^{\prime}\right)\right\|_{1}/\lambda)
\end{aligned}
\end{equation}

Consequently, we finish the proof of the node feature perturbation mechanism using the bound $S(f) \leq \lambda\epsilon_{x}$.

{\bf Weight Matrix Perturbation}. There are no second-order neighbors of nodes in graph $\mathcal{G}_{i}$, and all nodes are connected to other nodes only in neighboring layers. The adjacency matrix is only used to represent the structural information of the network, however, its structure is public and very sparse. To avoid too much extra computation, we focus privacy concerns on the pre-trained network's weights themselves. In this phase, we leverage a perturbation mechanism to protect its privacy for the weight matrix inspired by \cite{Graph_LDP}. We define the weight of each edge originating from the previous layer as the ``weight feature" of the nodes in the current layer. The ``weight feature" of the nodes shares the same dimension as the number of nodes in the previous layer.

As an illustration, take a model of ${U}_{i}$ that is fully connected. Given the current $p_{th}$ layer and its input weight list $\boldsymbol{w}_{p-1}$ from the previous $p-1_{th}$ layer, the ``weight feature" of the current layer is denoted as $\boldsymbol{w}_{i, p} = \boldsymbol{w}_{p-1} \in \mathbb{R}^{d}$, where $d$ is the number of nodes in the $p-1_{th}$ layer. In this way, we convert the perturbation of the weight list into a variant of private extraction of node features on the graph. In order to tackle the computational burden of perturbations, we perform a layer-wise perturbation by extending the multi-bit mechanism in \cite{LDPGNN} for private extraction of the ``weight feature" layer by layer. Assuming that each node of current $p_{th}$ layer owns $d$-dimensional ``weight feature" vector $\boldsymbol{w}_{i, p}$ whose elements fall into the range $ [\boldsymbol{w}_{min}, \boldsymbol{w}_{max}] $, we apply the perturbation on $\boldsymbol{w}_{i, p}$ layer by layer to get the corresponding perturbed ``weight feature" vector $\boldsymbol{\tilde{w}}_{i, p}$. Sharing the ``weight feature" list $\boldsymbol{\tilde{W}}_{i}$ with other model owners, after all of the ``weight feature" perturbations are completed. The process can be formulated as

\begin{equation}
\boldsymbol{\tilde{w}}_{i, p} = \boldsymbol{w}_{i, p} + \textit { MultiBit }\left( {\epsilon_{w}}\right)
\end{equation}
  
Instead of perturbing all the $d$ features, the perturbation mechanism uniformly samples $m$ out of $d$ features without replacement, which with a probability formulated as

\begin{equation}
\operatorname{Bernoulli}\left(\frac{1}{e^{\epsilon_{w} / m_{+1}}}+\frac{\boldsymbol{w}_{i, p}-\boldsymbol{w}_{min}}{\boldsymbol{w}_{max}-\boldsymbol{w}_{min}} \cdot \frac{e^{\epsilon_{w} / m_{-1}}}{e^{\epsilon_{w} / m_{+1}}}\right)
\end{equation}

For each sampled feature, a corresponding Bernoulli variable is drawn from the distribution whose parameter depends on the value of the feature and the privacy budget $\epsilon_{w}$. When $m = d$, the perturbation reduces to applying the 1-bit mechanism with a privacy budget of $\epsilon_{w}/d$ to every single feature. The following theorem ensures that the perturbation mechanism is $\epsilon_{w}$-LDP.

{\bf\textit{Theorem 2.}} \textit{The weight matrix perturbation mechanism $\mathcal{M}_{w}$ preserves $\epsilon_{w}$-local differential privacy.}

{\bf\textit{Proof.}} The perturbation mechanism $\mathcal{M}_{w}$ is based multi-bit mechanism applied on the ``weight feature" vector $\boldsymbol{w}_{i, p}$. According to the proof of \cite{LDPGNN}, we show that for any two weight features $\boldsymbol{w}_{i, p, 1}$ and $\boldsymbol{w}_{i, p, 2}$, we have $\frac{\boldsymbol{Pr}\left[\mathcal{M}_{w}(\boldsymbol{w}_{i, p, 1})=\boldsymbol{\tilde{w}}_{i, p}\right]}{\boldsymbol{Pr}\left[\mathcal{M}_{w}\left(\boldsymbol{w}_{i, p, 2}\right)=\boldsymbol{\tilde{w}}_{i, p}\right]} \leq e^{\epsilon_{w}}$. For any dimension $q \in \left\{1, 2, \ldots, d\right\}$, we get $\boldsymbol{\tilde{w}}_{i, p, q} \in \left\{-1, 0, 1\right\}$. When $q\notin \mathcal{S}$ with probability $1-m/d$, the $\boldsymbol{\tilde{w}}_{i, p, q} = 0$, therefore:

\begin{equation}
\frac{\boldsymbol{Pr}\left[\mathcal{M}_{w}(\boldsymbol{w}_{i, p, 1})_{q}=0\right]}{\boldsymbol{Pr}\left[\mathcal{M}_{w}(\boldsymbol{w}_{i, p, 2})_{q}=0\right]} = \frac{1-m/d}{1-m/d} = 1 \leq e^{\epsilon_{w}}, \forall \epsilon_{w} > 0
\end{equation}

Similarly, for $\boldsymbol{\tilde{w}}_{i, p} \in \left\{-1, 1\right\}$, we obtain the following inequality:

\begin{equation}
\frac{\boldsymbol{Pr}\left[\mathcal{M}_{w}(\boldsymbol{w}_{i, p, 1})_{q}\in \left\{-1, 1\right\}\right]}{\boldsymbol{Pr}\left[\mathcal{M}_{w}(\boldsymbol{w}_{i, p, 1})_{q}\in \left\{-1, 1\right\}\right]} \leq \frac{\frac{m}{d} \cdot \frac{e^{\epsilon_{w} / m_{-1}}}{e^{\epsilon_{w} / m_{+1}}}}{\frac{m}{d} \cdot \frac{1}{e^{\epsilon_{w} / m_{+1}}}} \leq e^{\epsilon_{w/m}}
\end{equation}

Therefore, with layer-by-layer iterations, we have
\begin{equation}
\begin{aligned}
&\frac{\boldsymbol{Pr}\left[\mathcal{M}_{w}\left(\boldsymbol{w}_{i, p, 1}\right)=\boldsymbol{\tilde{w}}_{i, p}\right]}{\boldsymbol{Pr}\left[\mathcal{M}_{w}\left(\boldsymbol{w}_{i, p, 2}\right)=\boldsymbol{\tilde{w}}_{i, p}\right]} = \prod_{q=1}^{d} \frac{\boldsymbol{Pr}\left[\mathcal{M}_{w}(\boldsymbol{w}_{i, p, 1})_{q}=\boldsymbol{\tilde{w}}_{i, p, q}\right]}{\boldsymbol{Pr}\left[\mathcal{M}_{w}(\boldsymbol{w}_{i, p, 2})_{q}=\boldsymbol{\tilde{w}}_{i, p, q}\right]} \\
& \quad\quad\quad = \prod_{y \mid \boldsymbol{\tilde{w}}_{i, p, y}=0} \frac{\boldsymbol{Pr}\left[\mathcal{M}_{w}\left(\boldsymbol{w}_{i, p, 1}\right)_{y}=0\right]}{\boldsymbol{Pr}\left[\mathcal{M}_{w}\left(\boldsymbol{w}_{i, p, 2}\right)_{y}=0\right]}  \\
& \quad\quad\quad \times \prod_{z \mid \boldsymbol{\tilde{w}}_{i, p, z} \in\{-1,1\}} \frac{\boldsymbol{Pr}\left[\mathcal{M}_{w}\left(\boldsymbol{w}_{i, p, 1}\right)_{z} \in\{-1,1\}\right]}{\boldsymbol{Pr}\left[\mathcal{M}_{w}\left(\boldsymbol{w}_{i, p, 2}\right)_{z} \in\{-1,1\}\right]} \\
& \quad\quad\quad = \prod_{\boldsymbol{\tilde{w}}_{i, p, z} \in\{-1,1\}} \frac{\boldsymbol{Pr}\left[\mathcal{M}_{w}\left(\boldsymbol{w}_{i, p, 1}\right)_{z} \in\{-1,1\}\right]}{\boldsymbol{Pr}\left[\mathcal{M}_{w}\left(\boldsymbol{w}_{i, p, 2}\right)_{z} \in\{-1,1\}\right]} \\
& \quad\quad\quad \leq \prod_{\boldsymbol{\tilde{w}}_{i, p, z} \in\{-1,1\}} e^{\epsilon_{w} / m} \\
& \quad\quad\quad \leq  e^{\epsilon_{w}}
\end{aligned}
\end{equation}

Thus, we complete the proof of the weight matrix perturbation mechanism.
 
\subsection{Decentralized Federated Graph Matching}
By performing information extraction and perturbation on all pre-trained models, the model owners can obtain privacy-preserving information that can be shared for cross-client model fusion. The goal is to find a suitable method that can achieve optimal fusion results without relying on the training data or revealing the original details of the models. One naive approach is to directly average the weights of corresponding positions in the weight space, similar to the FedAvg algorithm \cite{fedavg}. However, it's important to note that FedAvg still requires the participation of training data from each client during the aggregation process. While the idea of aggregation is similar, the proposed method aims to achieve model fusion without relying on the explicit use of training data. Moreover, recent popular studies \cite{OTfusion, GAMF, Git} have demonstrated the limitations of the above idea and utilized matching and alignment of network neurons for model fusion and effectively improved the accuracy of the fusion model. To address model privacy concerns, we propose a novel approach using decentralized federated graph matching. By leveraging a small amount of data for network inference and recording activation values, we preserve the privacy of the original training data while obtaining valuable information for fusion. This enables us to balance model performance and privacy preservation, facilitating effective model fusion with minimal data involvement.

In this process, the model owner ${U}_{i}$ shares the perturbed graph features $\boldsymbol{\tilde X}_{i}$ and $\boldsymbol{\tilde{W}}_{i}$ to the ${U}_{j}$ instead of directly sharing the true values of the model weights. The ${U}_{j}$ takes the received perturbed activation and weight of the model from ${U}_{i}$ to calculate the similarity with the neurons of the model from ${U}_{j}$ layer by layer. The similarity is denoted as an affinity matrix in graph matching, which can be used to solve the permutation matrix according to Eq.(1), to adjust the ordering of neurons, and then to fuse the model after the alignment of neurons. In our methodology, we evaluate the affinity of neuron activations and weights based on the \textit{Gaussian Kernel} applied by \cite{GAMF}, respectively. Then we merge these two matrices $\boldsymbol{\tilde{K_w}}$ and $\boldsymbol{\tilde{K_{act}}}$ together after normalization, as follows

\begin{equation}
\boldsymbol{\tilde{K}}_{[a, b, f, k]} = \boldsymbol{\tilde{K_w}}_{[a, b, f, k]} + \langle  \boldsymbol{\tilde{K_{act}}}_{[a, f]} \rangle_{s=(1, 1)} 
\end{equation}

Since the dimensions of the two matrices are different, we use a sliding window-like scheme $\langle \cdot \rangle$, with $strides=(1, 1)$, to merge them by adding them together and reducing the influence of the magnitude between the indicators by means of layer-wise normalization to ensure the efficiency of the merged affinity matrix.

Given the merged affinity matrix $\boldsymbol{\tilde{K}}_{[a, b, f, k]}$, the model alignment's target is to maximize the following objective function:

\begin{equation}
\max _{\boldsymbol{P}} \sum_{a=0}^{d_{\Sigma}-1} \sum_{b=0}^{d_{\Sigma}-1} \boldsymbol{R}_{[a, b]} \boldsymbol{P}_{[a, b]} \text { s.t. constraints in Eq. (1). }
\end{equation}
where $ \boldsymbol{R}_{[a, b]}=\sum_{f=0}^{d_{\Sigma}-1} \sum_{k=0}^{d_{\Sigma}-1} \boldsymbol{\tilde{K}}_{[a, b, f, k]} \boldsymbol{P}_{[f, k]}$. The constrained optimization problem described above is a linear assignment problem that usually is solved using the Hungarian algorithm \cite{Hungarian} or the Sinkhorn algorithm \cite{Sinkhorn} with a relaxed projection.

\subsection{Private Neuron Alignment and Model Fusion}
The problem of Eq.(15) is an NP-hard and a special case of \textit{Lawler’s Quadratic Assignment Problem} whose memory cost can be $O((d_{\Sigma})^{4})$ and difficult to avoid. To alleviate the computational complexity, we refer to the formulation of \cite{GAMF} and obtain the private permutation matrix $\boldsymbol{\tilde{P}}$ with decomposed diagonal matrices in a differentially private manner. Subsequently, we solved the objective function by the ``\textit{Spectral Graph Matching}" solver \cite{SMsolverl}, a method that obtains the eigenvectors of the input affinity matrix by power iteration. Thus, we can obtain the aligned model by permuting the node ordering using $\boldsymbol{\tilde{P}}$.

{\bf Private Model Fusion}. If we have two models with aligned neural networks, the challenge lies in effectively integrating them while maintaining privacy. One straightforward approach is to apply the LDP mechanism and aggregate the perturbed weights. However, this approach may lead to a cumulative degradation of model accuracy over time. Therefore, we design a \textit{Perturbation-Filter Adapter(PFA)} based on Gaussian randomization and search for a trade-off between privacy and accuracy for the exchange of model weights. The PFA consists of two parts, \textit{Randomized Perturbation Unit (RPU)} and \textit{Smoothing Filter Unit (SFU)}. Before sharing the weight of the aligned neural networks, the RPU performs perturbation on the weight $\boldsymbol{Wp}_{i}$, after the neurons have been permuted, by adding randomized Gaussian noise, that is

\begin{equation}
\widetilde{\boldsymbol{W}}_{i} \leftarrow \boldsymbol{\tilde{Wp}}_{i} = \boldsymbol{Wp}_{i} + \textit {Gaussian }\left({\epsilon_{f}, \delta}\right)
\end{equation}
where $\epsilon_{f}, \delta > 0$ and $\delta$ is typically small. According to definition in \cite{DP, DP_06}, the PRU satisfies the $(\epsilon_{f}, \delta)$-LDP. In, particular, $(\epsilon_{f}, \delta)$-LDP is also called approximate LDP and the case of $\delta = 0$ is called pure LDP. 

To reduce the performance degradation of model fusion, we use SFU to standardize the noise-added results to ensure the effectiveness of the model. Since the noise follows the Gaussian distribution, then we transform it through the corresponding probability density function. We compute the 
\textit{Gaussian Error Function} which denoted as $erf(\cdot)$, and could be standardized as following

\begin{equation}
\boldsymbol{W}_{i} = \max \left\{0, \operatorname{\textit{erf}}\left(\frac{\boldsymbol{\widetilde{\boldsymbol{W}}_{i}}-\mu}{\sigma \cdot \sqrt{2}}\right)\right\}
\end{equation}
where $\mu$ and $\delta$ denote the mean and standard deviation of the weights. Ultimately, after performing the above operations, we fuse the two models in the weight space by averaging them according to different scales $\alpha$ and save the model with the best performance.

In the model fusion phase, the ${U}_{i}$ fuses its model $\boldsymbol{W}_{i}$ with the model $\boldsymbol{W}_{j}$ at varying scales, allowing users to choose the fusion performance according to their specific needs. The fusion process is represented as follows

\begin{equation}
\boldsymbol{W}_{\mathcal{F}} = \frac{1}{2}\left(\alpha \times \boldsymbol{W}_{i} + (1 - \alpha) \times \boldsymbol{W}_{j}\right)    
\end{equation}

The securely fused model can then be reused and shared with other users for access and utilization.

{\bf\textit{Theorem 3.}} \textit{The PFA mechanism $\mathcal{M}_{f}$ preserves $(\epsilon_{f}, \delta)$-local differential privacy.}

{\bf\textit{Proof.}} Due to the page limitation, we refer to the detailed proof of Theorem 3 to \cite{DP_06} and \cite{DP}. The mechanism $\mathcal{M}_{f}$ achieves $(\epsilon_{f}, \delta)$-LDP with the probability at least $1 - \delta$. In \cite{DP_06}, the noise with standard deviation to each weight with $l_{2}$-sensitivity $\textit{sen}$ and was given by $\delta = \sqrt{2 \ln \frac{2}{\delta}} \cdot \frac{\textit{sen}}{\epsilon}$. Likewise, for delta in \cite{DP}, it was given by $\delta = \sqrt{2 \ln \frac{1.25}{\delta}} \cdot \frac{\textit{sen}}{\epsilon}$. Especially, the noise we added following a Gaussian distribution with mean 0 and variance $\delta ^2$.

{\bf\textit{Corollary.}} \textit{The randomized mechanisms $\mathcal{M}_{x}, \mathcal{M}_{w}$ and $\mathcal{M}_{f}$ jointly ensure $(\epsilon_{x} + \epsilon_{w} + \epsilon_{f})$-local differential privacy}. 

Therefore, the following process will also provide local differential privacy for the entire multi-model fusion and pre-trained model pool applications, referring to the composition theorem in \cite{DP}.

\section{Experiments}
We conduct a comprehensive set of experiments to evaluate the privacy-utility performance of the proposed method PrivFusion. Our investigation addresses various parameter settings that may impact its effectiveness, and aims to answer the following research questions:
\begin{itemize}
    \item \textbf{RQ1:} How does the performance of PrivFusion compare to conventional methods that have no privacy guarantees?
    \item \textbf{RQ2:} Is PrivFusion's performance controllable and does it achieve the desired utility-privacy trade-off under different privacy budgets?
    \item \textbf{RQ3:} Which factor, perturbing activation values or perturbing weights, plays a more critical role in the effectiveness of model fusion when adjusting the privacy budget? 
    \item \textbf{RQ4:} How does the hybrid differentially private randomized mechanism perform? Does it offer advantages over single mechanisms?
\end{itemize}

\begin{table*}[!t]
\caption{Comparison of two pre-trained models fusion on two public datasets}
\label{tab:table1}
\centering
\setlength{\tabcolsep}{8pt}
\begin{tabular}{@{}cccccccccc@{}}
\toprule
Dataset                                        & Data Partition                                      & Metric                        & \begin{tabular}[c]{@{}c@{}}Individual Models\\ {[}Acc(\%) , ...{]}\end{tabular} & \begin{tabular}[c]{@{}c@{}}PE\\ Acc(\%)\end{tabular} & \begin{tabular}[c]{@{}c@{}}VA\\ Acc(\%)\end{tabular} & \begin{tabular}[c]{@{}c@{}}OTFusion\\ Acc(\%)\end{tabular} & \begin{tabular}[c]{@{}c@{}}GAMF\\ Acc(\%)\end{tabular} & \begin{tabular}[c]{@{}c@{}}GIT \\ RE-BASIN\\ Acc(\%)\end{tabular} & \begin{tabular}[c]{@{}c@{}}PrivFusion\\ (ours)\\ Acc(\%)\end{tabular} \\ \midrule
\multicolumn{1}{c|}{\multirow{4}{*}{MNIST}}    & \multicolumn{1}{c|}{\multirow{2}{*}{Homogeneous}}   & \multicolumn{1}{c|}{best}     & \multicolumn{1}{c|}{\multirow{2}{*}{{[}93.06, 93.17{]}}}                        & \textbf{93.41}                                       & 92.85                                                & 93.13                                                      & 93.17                                                  & \multicolumn{1}{c|}{93.11}                                        & 93.01                                                                 \\
\multicolumn{1}{c|}{}                          & \multicolumn{1}{c|}{}                               & \multicolumn{1}{c|}{top3-Avg} & \multicolumn{1}{c|}{}                                                           & \textbf{93.39}                                       & 92.23                                                & 92.99                                                      & 93.17                                                  & \multicolumn{1}{c|}{93.05}                                        & 92.90                                                                 \\ \cmidrule(l){2-10} 
\multicolumn{1}{c|}{}                          & \multicolumn{1}{c|}{\multirow{2}{*}{Heterogeneous}} & \multicolumn{1}{c|}{best}     & \multicolumn{1}{c|}{\multirow{2}{*}{{[}77.63, 91.11{]}}}                        & 90.74                                                & 90.91                                                & 90.64                                                      & \textbf{91.11}                                         & \multicolumn{1}{c|}{77.63}                                        & 91.10                                                                 \\
\multicolumn{1}{c|}{}                          & \multicolumn{1}{c|}{}                               & \multicolumn{1}{c|}{top3-Avg} & \multicolumn{1}{c|}{}                                                           & 90.25                                                & 90.08                                                & 89.79                                                      & \textbf{91.10}                                         & \multicolumn{1}{c|}{77.46}                                        & 91.09                                                                 \\ \midrule
\multicolumn{1}{c|}{\multirow{4}{*}{CIFAR-10}} & \multicolumn{1}{c|}{\multirow{2}{*}{Homogeneous}}   & \multicolumn{1}{c|}{best}     & \multicolumn{1}{c|}{\multirow{2}{*}{{[}55.75, 56.49{]}}}                        & \textbf{56.80}                                       & 52.00                                                & 55.49                                                      & 56.49                                                  & \multicolumn{1}{c|}{55.75}                                        & 56.56                                                                 \\
\multicolumn{1}{c|}{}                          & \multicolumn{1}{c|}{}                               & \multicolumn{1}{c|}{top3-Avg} & \multicolumn{1}{c|}{}                                                           & \textbf{56.75}                                       & 47.11                                                & 53.81                                                      & 56.49                                                  & \multicolumn{1}{c|}{53.98}                                        & 56.31                                                                 \\ \cmidrule(l){2-10} 
\multicolumn{1}{c|}{}                          & \multicolumn{1}{c|}{\multirow{2}{*}{Heterogeneous}} & \multicolumn{1}{c|}{best}     & \multicolumn{1}{c|}{\multirow{2}{*}{{[}44.06, 64.05{]}}}                        & 63.65                                                & 62.80                                                & 62.68                                                      & 64.08                                                  & \multicolumn{1}{c|}{46.43}                                        & \textbf{64.30}                                                        \\
\multicolumn{1}{c|}{}                          & \multicolumn{1}{c|}{}                               & \multicolumn{1}{c|}{top3-Avg} & \multicolumn{1}{c|}{}                                                           & 62.59                                                & 54.71                                                & 59.75                                                      & 64.06                                                  & \multicolumn{1}{c|}{46.11}                                        & \textbf{64.07}                                                        \\ \bottomrule
\end{tabular}
\end{table*}

\subsection{Experimental settings} 
\subsubsection{Datasets} 
In this section, we draw inspiration from two existing model fusion methods, namely OTFusion \cite{OTfusion} and GAMF \cite{GAMF}. We evaluate our proposed approach using the widely used image classification datasets, MNIST and CIFAR-10. The data augmentation settings employed in our experiments are consistent with those specified in OTFusion. For more comprehensive information about the datasets and augmentation settings, we recommend referring to their respective open-source repositories.

We explore two different data partition settings for local models/users: 1) Homogeneous data partition, where each client acquires data from an independent and identically distributed (IID) source, and the data distribution across classes is nearly equal among users. 2) Heterogeneous data partition, where each user acquires data from a non-IID source, resulting in variations in the data distribution across users. To simulate a heterogeneous data split, we followed the approach used in previous works such as \cite{OTfusion}. Specifically, for the MNIST digit classification task, we designed Model A to possess a unique capability of recognizing a specific ``personalized" label, which in this case was digit 4. Model B, on the other hand, comprised the majority of the remaining training set (excluding label 4), accounting for four-fifths of the data, while Model A held the remaining one-fifth. By incorporating these distinct data partition settings, we aim to evaluate the performance and robustness of our proposed method in scenarios where clients have varying access to different classes of data.

\subsubsection{Backbone Models and Training Settings} 
The graph-based model fusion approach we employed in our study has a memory allocation requirement for initialization, resulting in significant overhead. To accommodate these constraints within our experimental setup, we conducted evaluations using relatively modest network architectures. For the MNIST dataset, we selected a fully connected network comprising five layers, along with a small convolutional neural network (CNN) consisting of two convolutional layers and one fully connected layer. As for the CIFAR-10 dataset, we employed a classic CNN architecture with two convolutional layers and three fully connected layers. The training process involved training both models on the MNIST dataset for 20 epochs, while the CIFAR-10 training was conducted for a total of 200 epochs, with all other training settings kept consistent. 

By employing this standardized approach, we ensured fair and consistent comparisons between the compared methods when evaluating the two models across different datasets. While it is important to note that the scalability of our findings to larger-scale networks may be limited, our study provides valuable insights into the functionality and effectiveness of the proposed technique. Furthermore, \textbf{the PrivFusion can be easily scaled to larger networks with increased computational resources, offering the potential for future research in the context of larger-scale networks}.

\subsubsection{Compared Methods} 
We compare the model fusion performance of the proposed PrivFusion against the following ones:
\begin{itemize}

    \item \textbf{Prediction Ensemble (PE)}. Prediction ensembling involves retaining all models and averaging their predictions or output layer scores, representing the optimal performance that can be achieved by consolidating into a single model, albeit an unattainable one.
    
    \item \textbf{Vanilla Averaging (VA)} is a simple but efficient algorithm that denotes the direct averaging of parameters. This approach shares a similar concept with the classic method \textbf{FedAvg} \cite{fedavg} in the context of federated learning. Both methods aim to aggregate model parameters from different clients to obtain a global model that represents the collective knowledge of all participants.  

    \item \textbf{OTFusion} \cite{OTfusion}. As previously mentioned, published at \textit{NeurIPS 2020}, the model fusion problem is formulated as a linear assignment problem and solved using the Wasserstein barycenter.

    \item \textbf{GAMF} \cite{GAMF}. As aforementioned, published at \textit{ICML 2022}, it formulates the model fusion problem as a graph-matching task to maximize weight similarity and iteratively updates matchings in a consistency-maintaining manner.

    \item \textbf{GIT RE-BASIN} \cite{Git}. A prominent work, published at \textit{ICLR 2023}, which merges two models in weight space by introducing three algorithms that permute the units of one model to align with a reference model.

\end{itemize}

Due to the differences in the settings between our method and federated learning, we do not include comparisons with related works in the field of federated learning. In terms of privacy protection mechanisms, we have selected several independent mechanisms to compare with our proposed hybrid mechanism.
\begin{itemize}
    \item \textbf{Laplace Mechanism} \cite{Laplace_LDP}. The Laplace mechanism is commonly used for count queries, sum queries, and other low-sensitivity queries. It is particularly suitable for discrete and bounded data domains.
    
    \item \textbf{Gaussian Mechanism} \cite{DP}. The Gaussian mechanism is commonly employed for query tasks that involve continuous data domains. It is particularly suitable for high-sensitivity queries such as range queries and mean queries.
    
    \item \textbf{MultiBit Mechanism} \cite{LDPGNN}. The MultiBit mechanism, introduced at \textit{CCS 2021}, extends the 1-bit mechanism \cite{MSepsilon} for collecting multidimensional features. It has been shown to outperform the widely used mechanism in terms of performance.
\end{itemize}
\subsubsection{Experimental Setup} 
Our experiments were conducted on a computing platform comprising two Intel Xeon Silver 4214R processors, each with 12 cores and a total of 192GB of RAM, and two GTX 3090 GPUs. Following the experimental phase, we proceeded to conduct model fusion using various proportions for comparative analysis. The fusion involved combining the two models in different ratios from $\{0, 0.1, 0.2, ..., 1\}$, and their performance was assessed using the accuracy metric. 

To assess the impact of different fusion ratios on overall performance, we conducted a comparative analysis between the average results of the top-3 models and the best-performing model. This approach provided valuable insights into the effectiveness of diverse fusion strategies in enhancing accuracy. By considering both the average results of the top-3 models, which represent the stability of the fused models and the best-performing model, which represents the optimal performance available for users to choose from, we gained a comprehensive understanding of the influence of fusion ratios on performance enhancement.

\begin{table*}[]
\caption{Comparison of different privacy budgets' effect}
\label{tab:table2}
\centering
\setlength{\tabcolsep}{14pt}
\begin{tabular}{@{}cccccccc@{}}
\toprule
                                                 &                                                      &                                                                                                                & \multicolumn{3}{c}{Privacy Budget}                                                                              & \multicolumn{2}{c}{Acc(\%)}                                       \\
\multirow{-2}{*}{Dataset}                        & \multirow{-2}{*}{Data Partition}                     & \multirow{-2}{*}{\begin{tabular}[c]{@{}c@{}}Individual Models\\ {[}Acc(\%) of each model, ...{]}\end{tabular}} & $\epsilon_w$                            & $\epsilon_a$                            & $\epsilon_f$                                                 & top3-Avg                         & best                          \\ \midrule
\multicolumn{1}{c|}{}                            & \multicolumn{1}{c|}{}                                & \multicolumn{1}{c|}{}                                                                                          & \cellcolor[HTML]{EFEFEF}0.01 & \cellcolor[HTML]{EFEFEF}0.01 & \multicolumn{1}{c|}{\cellcolor[HTML]{EFEFEF}0.1}  & \cellcolor[HTML]{EFEFEF}87.50 $\pm$ 0.4 & \cellcolor[HTML]{EFEFEF}92.27 \\
\multicolumn{1}{c|}{}                            & \multicolumn{1}{c|}{}                                & \multicolumn{1}{c|}{}                                                                                          & 0.01                         & 0.1                          & \multicolumn{1}{c|}{0.1}                          & \textbf{90.48 $\pm$ 0.2}                & \textbf{92.66}                \\
\multicolumn{1}{c|}{}                            & \multicolumn{1}{c|}{}                                & \multicolumn{1}{c|}{}                                                                                          & 0.1                          & 0.01                         & \multicolumn{1}{c|}{0.1}                          & 88.29 $\pm$ 0.3                         & 92.41                         \\ \cmidrule(l){4-8} 
\multicolumn{1}{c|}{}                            & \multicolumn{1}{c|}{}                                & \multicolumn{1}{c|}{}                                                                                          & \cellcolor[HTML]{EFEFEF}0.01 & \cellcolor[HTML]{EFEFEF}0.01 & \multicolumn{1}{c|}{\cellcolor[HTML]{EFEFEF}0.01} & \cellcolor[HTML]{EFEFEF}19.97 $\pm$ 0.8 & \cellcolor[HTML]{EFEFEF}27.55 \\
\multicolumn{1}{c|}{}                            & \multicolumn{1}{c|}{}                                & \multicolumn{1}{c|}{}                                                                                          & 0.01                         & 0.1                          & \multicolumn{1}{c|}{0.01}                         & \textbf{24.58 $\pm$ 0.5}                & 33.12                         \\
\multicolumn{1}{c|}{}                            & \multicolumn{1}{c|}{\multirow{-6}{*}{Homogeneous}}   & \multicolumn{1}{c|}{\multirow{-6}{*}{{[}93.06, 93.17{]}}}                                                      & 0.1                          & 0.01                         & \multicolumn{1}{c|}{0.01}                         & 22.32 $\pm$ 0.7                         & \textbf{44.02}                \\ \cmidrule(l){2-8} 
\multicolumn{1}{c|}{}                            & \multicolumn{1}{c|}{}                                & \multicolumn{1}{c|}{}                                                                                          & \cellcolor[HTML]{EFEFEF}0.01 & \cellcolor[HTML]{EFEFEF}0.01 & \multicolumn{1}{c|}{\cellcolor[HTML]{EFEFEF}0.1}  & \cellcolor[HTML]{EFEFEF}89.41 $\pm$ 0.2 & \cellcolor[HTML]{EFEFEF}90.91 \\
\multicolumn{1}{c|}{}                            & \multicolumn{1}{c|}{}                                & \multicolumn{1}{c|}{}                                                                                          & 0.01                         & 0.1                          & \multicolumn{1}{c|}{0.1}                          & \textbf{89.67 $\pm$ 0.1}                & 90.83                         \\
\multicolumn{1}{c|}{}                            & \multicolumn{1}{c|}{}                                & \multicolumn{1}{c|}{}                                                                                          & 0.1                          & 0.01                         & \multicolumn{1}{c|}{0.1}                          & 89.47 $\pm$ 0.3                         & \textbf{91.02}                \\ \cmidrule(l){4-8} 
\multicolumn{1}{c|}{}                            & \multicolumn{1}{c|}{}                                & \multicolumn{1}{c|}{}                                                                                          & \cellcolor[HTML]{EFEFEF}0.01 & \cellcolor[HTML]{EFEFEF}0.01 & \multicolumn{1}{c|}{\cellcolor[HTML]{EFEFEF}0.01} & \cellcolor[HTML]{EFEFEF}37.80 $\pm$ 0.7 & \cellcolor[HTML]{EFEFEF}41.86 \\
\multicolumn{1}{c|}{}                            & \multicolumn{1}{c|}{}                                & \multicolumn{1}{c|}{}                                                                                          & 0.01                         & 0.1                          & \multicolumn{1}{c|}{0.01}                         & \textbf{46.49 $\pm$ 0.4}                & 66.00                         \\
\multicolumn{1}{c|}{\multirow{-12}{*}{MNIST}}    & \multicolumn{1}{c|}{\multirow{-6}{*}{Heterogeneous}} & \multicolumn{1}{c|}{\multirow{-6}{*}{{[}77.63, 91.11{]}}}                                                      & 0.1                          & 0.01                         & \multicolumn{1}{c|}{0.01}                         & 43.61 $\pm$ 0.5                         & \textbf{72.78}                \\ \midrule
\multicolumn{1}{c|}{}                            & \multicolumn{1}{c|}{}                                & \multicolumn{1}{c|}{}                                                                                          & \cellcolor[HTML]{EFEFEF}0.01 & \cellcolor[HTML]{EFEFEF}0.01 & \multicolumn{1}{c|}{\cellcolor[HTML]{EFEFEF}0.1}  & \cellcolor[HTML]{EFEFEF}44.11 $\pm$ 0.3 & \cellcolor[HTML]{EFEFEF}52.05 \\
\multicolumn{1}{c|}{}                            & \multicolumn{1}{c|}{}                                & \multicolumn{1}{c|}{}                                                                                          & 0.01                         & 0.1                          & \multicolumn{1}{c|}{0.1}                          & \textbf{47.15 $\pm$ 0.3}                & \textbf{54.04}                \\
\multicolumn{1}{c|}{}                            & \multicolumn{1}{c|}{}                                & \multicolumn{1}{c|}{}                                                                                          & 0.1                          & 0.01                         & \multicolumn{1}{c|}{0.1}                          & 45.02 $\pm$ 0.5                         & 53.83                         \\ \cmidrule(l){4-8} 
\multicolumn{1}{c|}{}                            & \multicolumn{1}{c|}{}                                & \multicolumn{1}{c|}{}                                                                                          & \cellcolor[HTML]{EFEFEF}0.01 & \cellcolor[HTML]{EFEFEF}0.01 & \multicolumn{1}{c|}{\cellcolor[HTML]{EFEFEF}0.01} & \cellcolor[HTML]{EFEFEF}14.75 $\pm$ 0.8 & \cellcolor[HTML]{EFEFEF}16.22 \\
\multicolumn{1}{c|}{}                            & \multicolumn{1}{c|}{}                                & \multicolumn{1}{c|}{}                                                                                          & 0.01                         & 0.1                          & \multicolumn{1}{c|}{0.01}                         & \textbf{15.39 $\pm$ 0.5}                & \textbf{18.54}                \\
\multicolumn{1}{c|}{}                            & \multicolumn{1}{c|}{\multirow{-6}{*}{Homogeneous}}   & \multicolumn{1}{c|}{\multirow{-6}{*}{{[}55.75, 56.49{]}}}                                                      & 0.1                          & 0.01                         & \multicolumn{1}{c|}{0.01}                         & 14.84 $\pm$ 0.4                         & 17.47                         \\ \cmidrule(l){2-8} 
\multicolumn{1}{c|}{}                            & \multicolumn{1}{c|}{}                                & \multicolumn{1}{c|}{}                                                                                          & \cellcolor[HTML]{EFEFEF}0.01 & \cellcolor[HTML]{EFEFEF}0.01 & \multicolumn{1}{c|}{\cellcolor[HTML]{EFEFEF}0.1}  & \cellcolor[HTML]{EFEFEF}51.96 $\pm$ 0.4 & \cellcolor[HTML]{EFEFEF}62.48 \\
\multicolumn{1}{c|}{}                            & \multicolumn{1}{c|}{}                                & \multicolumn{1}{c|}{}                                                                                          & 0.01                         & 0.1                          & \multicolumn{1}{c|}{0.1}                          & \textbf{55.13 $\pm$ 0.2}                & \textbf{63.63}                \\
\multicolumn{1}{c|}{}                            & \multicolumn{1}{c|}{}                                & \multicolumn{1}{c|}{}                                                                                          & 0.1                          & 0.01                         & \multicolumn{1}{c|}{0.1}                          & 52.66 $\pm$ 0.6                         & 63.57                         \\ \cmidrule(l){4-8} 
\multicolumn{1}{c|}{}                            & \multicolumn{1}{c|}{}                                & \multicolumn{1}{c|}{}                                                                                          & \cellcolor[HTML]{EFEFEF}0.01 & \cellcolor[HTML]{EFEFEF}0.01 & \multicolumn{1}{c|}{\cellcolor[HTML]{EFEFEF}0.01} & \cellcolor[HTML]{EFEFEF}14.85 $\pm$ 0.7 & \cellcolor[HTML]{EFEFEF}20.01 \\
\multicolumn{1}{c|}{}                            & \multicolumn{1}{c|}{}                                & \multicolumn{1}{c|}{}                                                                                          & 0.01                         & 0.1                          & \multicolumn{1}{c|}{0.01}                         & \textbf{17.45 $\pm$ 0.5}                & \textbf{23.02}                \\
\multicolumn{1}{c|}{\multirow{-12}{*}{CIFAR-10}} & \multicolumn{1}{c|}{\multirow{-6}{*}{Heterogeneous}} & \multicolumn{1}{c|}{\multirow{-6}{*}{{[}44.06, 64.05{]}}}                                                      & 0.1                          & 0.01                         & \multicolumn{1}{c|}{0.01}                         & 16.77 $\pm$ 0.7                         & 22.06         \\ \bottomrule               
\end{tabular}
\end{table*}

\subsection{Performance Evaluation}
\textbf{\textit{RQ1: PrivFusion demonstrates high utility accompanied by acceptable performance loss}}. Table I presents a comprehensive performance analysis of different methods on the MNIST and CIFAR-10 datasets. Two data partitioning approaches, homogeneous and heterogeneous, are considered for evaluation. In this study, our proposed method focuses on striking a balance between privacy protection and utility. Across the MNIST dataset, the PrivFusion, with all privacy budgets consistently set to 1, achieves performance that is comparable to the optimal method under both data partitioning approaches. This indicates that our approach effectively manages the trade-off between preserving privacy and maintaining model accuracy. Specifically, when using the homogeneous data partition, our method demonstrates competitive results compared to PE, VA, GAMF, OTFusion and GIT RE-BASIN. Moreover, when considering the average of the top 3 model accuracy and the best model accuracy, the PrivFusion performs comparably to the optimal method (highlighted in bold) under both data partitioning approaches.

Moving to the CIFAR-10 dataset, the PrivFusion delivers impressive results by achieving high model accuracy while preserving privacy. In the Homogeneous data partition scenario, our method outperforms VA and OTFusion, while achieving comparable results to PE, GIT RE-BASIN and GAMF. The PrivFusion consistently exhibits competitive performance on the CIPAF-10 compared to all other methods being evaluated. Especially in the case of heterogeneous data partitioning, PrivFusion achieves the best performance, which may be attributed to the unpredictable contribution of randomized perturbations to the models. These findings underscore the practical potential of our approach in privacy-preserving machine learning applications.

\begin{figure}[t]
\includegraphics[width=\linewidth]{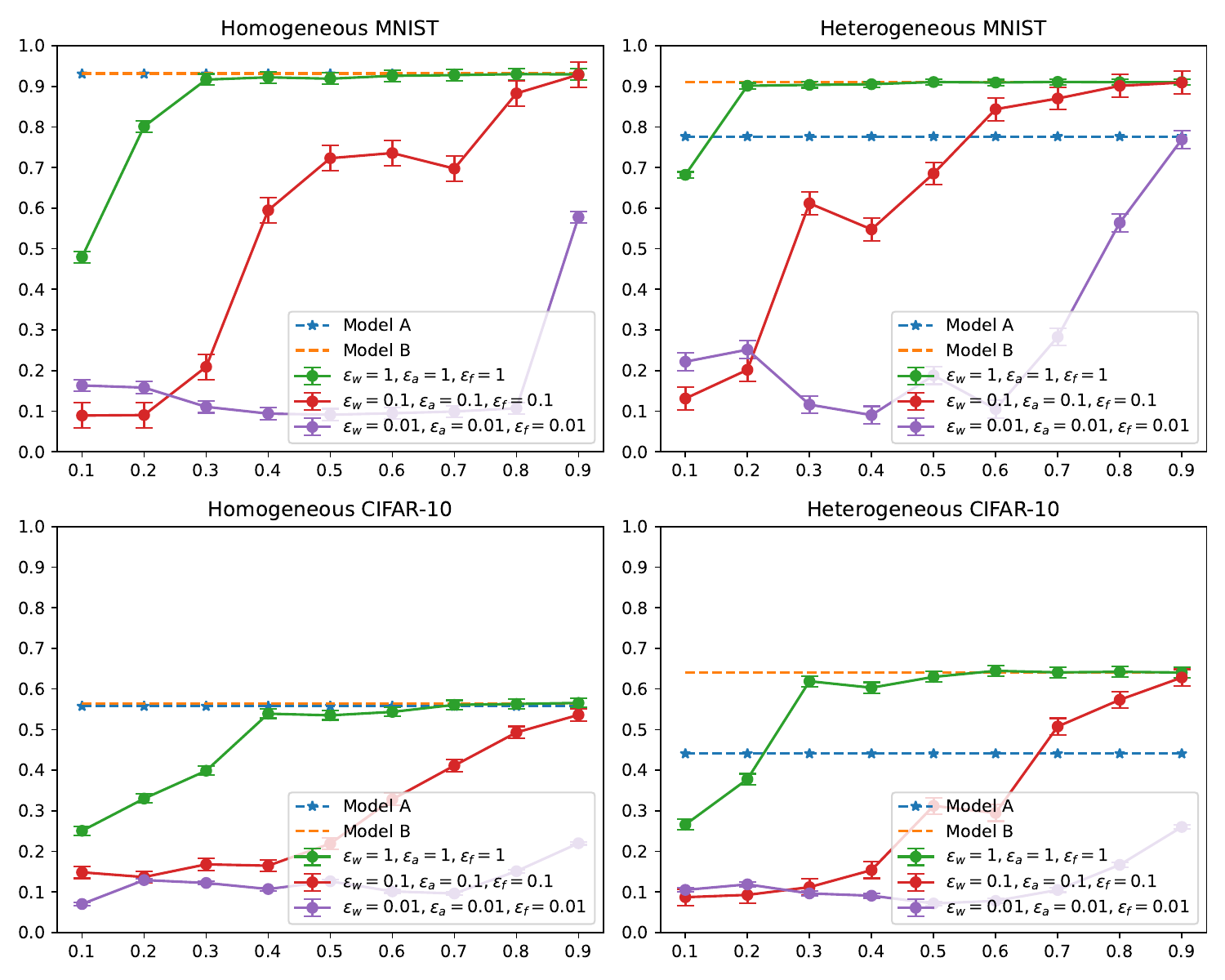}
\caption{The effect of the privacy budget on the performance of PrivFusion.}
\label{fig:ex2}
\end{figure}

\textbf{\textit{RQ2: The perturbation of PrivFusion is effective and controllable, allowing for the achievement of a utility-privacy trade-off by adjusting the privacy budget}}. Fig. 3 illustrates the influence of adjusting the privacy budget on the accuracy of the fused model, highlighting the controllable nature of this adjustment. By manipulating the privacy budget, the degree of perturbation applied to the model can be controlled, leading to different levels of performance degradation. When the privacy budget is set to 1, indicating minimal perturbation, the fused model achieves the highest accuracy among different privacy budget settings. This implies that with a larger privacy budget and smaller perturbations, the model retains a higher level of accuracy. As the privacy budget decreases to 0.1, indicating increased perturbations, there is a slight reduction in model accuracy compared to the budget of 1. However, the observed performance degradation remains relatively small, suggesting that the model can tolerate a certain level of perturbation while maintaining reasonable accuracy. Further reducing the privacy budget to 0.01, representing even larger perturbations, leads to a more pronounced decline in model accuracy. The increased perturbation significantly impacts the model's performance, resulting in a noticeable reduction in accuracy.

These findings highlight the fundamental trade-off between privacy preservation and model performance. By adjusting the privacy budget, practitioners can exert control over the level of perturbation applied to the model, allowing them to influence the delicate balance between maintaining privacy and preserving model accuracy. This control over the privacy-performance trade-off provides valuable flexibility for practitioners in privacy-preserving machine learning scenarios.

\textbf{\textit{RQ3: The perturbation of activations has a greater influence compared to weights on fused model performance}}. Table II illustrates the influence of adjusting the privacy budget, which corresponds to varying levels of perturbation strength, on the perturbation of model weights($\epsilon_w$), activation values($\epsilon_a$), and the private fusion($\epsilon_f$) process. The results highlight the importance of the privacy budget in determining the extent of perturbation and its impact on the final results. Specifically, the fused model can get the optimal performance (highlighted in bold with their standard deviation) when $\epsilon_a > \epsilon_w$, and the analysis reveals that perturbing activation values have a greater impact on the final results compared to perturbing model weights. This suggests that privacy-preserving techniques should focus more on perturbing activation values while minimizing the perturbation on model weights to achieve better results. Additionally, it is observed that the perturbation strength in fusion, the $\epsilon_f$ which setting to 0.1 and 0.01, has the greatest impact on the accuracy of the fused model. This parameter is especially crucial when performing model exchange and sharing. Furthermore, the analysis shows that perturbing activation values and model weights mainly affect the performance of neuron similarity and matching mapping. This suggests that perturbing these factors may not significantly impact the overall performance of the model.

In conclusion, the results demonstrate that adjusting the privacy budget plays a crucial role in privacy-preserving machine learning. Focusing on perturbing activation values and minimizing the perturbation on model weights can lead to improved results. Moreover, the perturbation strength in fusion is a crucial parameter that significantly impacts the accuracy of the fused model. The findings can inform the development of privacy-preserving techniques for machine learning and contribute to the advancement of the field.

\textbf{\textit{RQ4: The hybrid differentially private randomized mechanism is more effective compared to a single mechanism}}. In Table III, we present the results which validate the superiority of the hybrid mechanism in model fusion performance. The comparative methods include the Non-Private mechanism, Laplace mechanism, Gaussian mechanism, MultiBit mechanism, and our proposed hybrid mechanism with Perturbation-Filter Adapter (PFA). The non-private mechanism serves as the upper bound for model fusion performance, while other methods exhibit performance degradation to varying degrees. Each method employs a specific perturbation mechanism in different stages of the fusion process. The results unequivocally demonstrate that the hybrid mechanism outperforms the other methods in terms of model accuracy.

For different components, it is recommended to apply different perturbation mechanisms. Specifically, \textbf{the Laplace mechanism with lower sensitivity is suitable for perturbing activation values, as it provides strong differential privacy guarantees and enhances security. On the other hand, the MultiBit mechanism is effective for perturbing weights, as it ensures both utility and tolerance for model parameter count and precision bits. Lastly, for the final private model fusion, the Gaussian mechanism based on a uniform distribution with a more relaxed privacy guarantee can be utilized. This hybrid mechanism achieves a balanced trade-off between privacy preservation and model utility.} These findings align with the theoretical analysis, which rigorously demonstrates that the hybrid mechanism satisfies the constraints and requirements of local differential privacy.

\begin{table}[]
\caption{Comparison of different perturbation mechanisms effect}
\label{tab:table3}
\centering
\setlength{\tabcolsep}{26pt}
\begin{tabular}{@{}ccc@{}}
\toprule
\multirow{2}{*}{Perturbation Mechanism} & \multicolumn{2}{c}{   
        Acc(\%)}         \\
                                        & top3-Avg          & best           \\ \midrule
Non-Private                            & 93.17              & 93.17          \\
Laplace                                 & 87.14 $\pm$ 0.3          & 87.77          \\
Gaussian                                & 85.29 $\pm$ 0.5          & 87.31          \\
MultiBit                                & 90.87 $\pm$ 0.1          & 91.83          \\ \midrule
Hybrid (ours)                           & 88.81 $\pm$ 0.5          & 89.65          \\
Hybrid + PFA (ours)                     & \textbf{92.90 $\pm$ 0.2} & \textbf{93.01} \\ \bottomrule
\end{tabular}
\end{table}

Furthermore, integrating PFA with the hybrid mechanism enhances its performance even further. The experimental results corroborate the effectiveness of this integration, revealing improved model accuracy while ensuring privacy preservation. Notably, the performance of the hybrid mechanism with PFA closely approaches that of the non-private model fusion, indicating a minimal loss in model accuracy due to privacy protection. In summary, the analysis of Table III unequivocally confirms the superiority of the hybrid mechanism in model perturbation. It effectively addresses various perturbation tasks and strikes a delicate balance between privacy and utility, as evidenced by its superior model accuracy compared to single Laplace, Gaussian, and MultiBit mechanisms. The incorporation of the Perturbation-Filter Adapter (PFA) serves to augment the performance of the hybrid mechanism, bringing its fusion model closer to the lossless model fusion without privacy protection.

\begin{table*}[]
\caption{The performance of real-world application tasks with different metrics.}
\label{tab:table4}
\centering
\setlength{\tabcolsep}{10pt}
\begin{tabular}{@{}cccccccccc@{}}
\toprule
{\color[HTML]{333333} }                                                  & {\color[HTML]{333333} Models}                                             & {\color[HTML]{333333} \begin{tabular}[c]{@{}c@{}}Privacy Budget\\ {[}$\epsilon_w$, $\epsilon_a$, $\epsilon_f${]}\end{tabular}} & {\color[HTML]{333333} Acc}              & {\color[HTML]{333333} ma\_F1}            & {\color[HTML]{333333} w\_F1}             & {\color[HTML]{333333} ma\_Rec}            & {\color[HTML]{333333} w\_Rec}             & {\color[HTML]{333333} ma\_Prec}           & {\color[HTML]{333333} w\_Prec}            \\ \midrule
\multicolumn{1}{c|}{{\color[HTML]{333333} }}                             & \multicolumn{1}{c|}{{\color[HTML]{333333} Model A}}                       & \multicolumn{1}{c|}{{\color[HTML]{333333} -}}                                               & {\color[HTML]{333333} 77.05\%}          & {\color[HTML]{333333} 0.7677}          & {\color[HTML]{333333} 0.772}           & {\color[HTML]{333333} 77.39\%}          & {\color[HTML]{333333} 77.05\%}          & {\color[HTML]{333333} 76.69\%}          & {\color[HTML]{333333} 77.87\%}          \\
\multicolumn{1}{c|}{{\color[HTML]{333333} }}                             & \multicolumn{1}{c|}{{\color[HTML]{333333} Model B}}                       & \multicolumn{1}{c|}{{\color[HTML]{333333} -}}                                               & {\color[HTML]{333333} 77.30\%}          & {\color[HTML]{333333} 0.7653}          & {\color[HTML]{333333} 0.7725}          & {\color[HTML]{333333} 76.42\%}          & {\color[HTML]{333333} 77.30\%}          & {\color[HTML]{333333} 76.67\%}          & {\color[HTML]{333333} 77.22\%}          \\
\multicolumn{1}{c|}{{\color[HTML]{333333} }}                             & \multicolumn{1}{c|}{{\color[HTML]{333333} Fused Model}}                   & \multicolumn{1}{c|}{{\color[HTML]{333333} -}}                                               & {\color[HTML]{333333} 76.91\%}          & {\color[HTML]{333333} 0.7573}          & {\color[HTML]{333333} 0.7663}          & {\color[HTML]{333333} 75.34\%}          & {\color[HTML]{333333} 76.91\%}          & {\color[HTML]{333333} 76.59\%}          & {\color[HTML]{333333} 76.79\%}          \\ \cmidrule(l){2-10} 
\multicolumn{1}{c|}{{\color[HTML]{333333} }}                             & \multicolumn{1}{c|}{{\color[HTML]{333333} }}                              & \multicolumn{1}{c|}{{\color[HTML]{333333} {[}0.01, 0.1, 1{]}}}                              & {\color[HTML]{333333} \textbf{77.10\%}} & {\color[HTML]{333333} \textbf{0.7605}} & {\color[HTML]{333333} \textbf{0.769}}  & {\color[HTML]{333333} \textbf{75.72\%}} & {\color[HTML]{333333} \textbf{77.10\%}} & {\color[HTML]{333333} \textbf{76.67\%}} & {\color[HTML]{333333} \textbf{76.96\%}} \\
\multicolumn{1}{c|}{{\color[HTML]{333333} }}                             & \multicolumn{1}{c|}{{\color[HTML]{333333} }}                              & \multicolumn{1}{c|}{{\color[HTML]{333333} {[}0.01, 0.1, 0.1{]}}}                            & {\color[HTML]{333333} 76.57\%}          & {\color[HTML]{333333} 0.7539}          & {\color[HTML]{333333} 0.7629}          & {\color[HTML]{333333} 75.01\%}          & {\color[HTML]{333333} 76.57\%}          & {\color[HTML]{333333} 76.20\%}          & {\color[HTML]{333333} 76.43\%}          \\
\multicolumn{1}{c|}{\multirow{-6}{*}{{\color[HTML]{333333} unbalanced}}} & \multicolumn{1}{c|}{\multirow{-3}{*}{{\color[HTML]{333333} PrivFusion(ours)}}} & \multicolumn{1}{c|}{{\color[HTML]{333333} {[}0.01, 0.1, 0.01{]}}}                           & {\color[HTML]{333333} 63.65\%}          & {\color[HTML]{333333} 0.6243}          & {\color[HTML]{333333} 0.6357}          & {\color[HTML]{333333} 62.38\%}          & {\color[HTML]{333333} 63.65\%}          & {\color[HTML]{333333} 62.49\%}          & {\color[HTML]{333333} 63.50\%}          \\ \midrule
\multicolumn{1}{c|}{{\color[HTML]{333333} }}                             & \multicolumn{1}{c|}{{\color[HTML]{333333} Model A}}                       & \multicolumn{1}{c|}{{\color[HTML]{333333} -}}                                               & {\color[HTML]{333333} 61.94\%}          & {\color[HTML]{333333} 0.6191}          & {\color[HTML]{333333} 0.6173}          & {\color[HTML]{333333} 64.22\%}          & {\color[HTML]{333333} 61.94\%}          & {\color[HTML]{333333} 64.54\%}          & {\color[HTML]{333333} 66.50\%}          \\
\multicolumn{1}{c|}{{\color[HTML]{333333} }}                             & \multicolumn{1}{c|}{{\color[HTML]{333333} Model B}}                       & \multicolumn{1}{c|}{{\color[HTML]{333333} -}}                                               & {\color[HTML]{333333} 79.45\%}          & {\color[HTML]{333333} 0.7905}          & {\color[HTML]{333333} 0.7954}          & {\color[HTML]{333333} 79.36\%}          & {\color[HTML]{333333} 79.45\%}          & {\color[HTML]{333333} 78.88\%}          & {\color[HTML]{333333} 79.76\%}          \\
\multicolumn{1}{c|}{{\color[HTML]{333333} }}                             & \multicolumn{1}{c|}{{\color[HTML]{333333} Fused Model}}                   & \multicolumn{1}{c|}{{\color[HTML]{333333} -}}                                               & {\color[HTML]{333333} 80.14\%}          & {\color[HTML]{333333} 0.7936}          & {\color[HTML]{333333} 0.8004}          & {\color[HTML]{333333} 79.12\%}          & {\color[HTML]{333333} 80.14\%}          & {\color[HTML]{333333} 79.72\%}          & {\color[HTML]{333333} 80.04\%}          \\ \cmidrule(l){2-10} 
\multicolumn{1}{c|}{{\color[HTML]{333333} }}                             & \multicolumn{1}{c|}{{\color[HTML]{333333} }}                              & \multicolumn{1}{c|}{{\color[HTML]{333333} {[}0.01, 0.1, 1{]}}}                              & {\color[HTML]{333333} \textbf{80.14\%}} & {\color[HTML]{333333} \textbf{0.7941}} & {\color[HTML]{333333} \textbf{0.8006}} & {\color[HTML]{333333} \textbf{79.22\%}} & {\color[HTML]{333333} \textbf{80.14\%}} & {\color[HTML]{333333} \textbf{79.77\%}} & {\color[HTML]{333333} \textbf{80.05\%}} \\
\multicolumn{1}{c|}{{\color[HTML]{333333} }}                             & \multicolumn{1}{c|}{{\color[HTML]{333333} }}                              & \multicolumn{1}{c|}{{\color[HTML]{333333} {[}0.01, 0.1, 0.1{]}}}                            & {\color[HTML]{333333} 79.40\%}          & {\color[HTML]{333333} 0.7862}          & {\color[HTML]{333333} 0.7931}          & {\color[HTML]{333333} 78.40\%}          & {\color[HTML]{333333} 79.40\%}          & {\color[HTML]{333333} 78.93\%}          & {\color[HTML]{333333} 79.30\%}          \\
\multicolumn{1}{c|}{\multirow{-6}{*}{{\color[HTML]{333333} sd2t2}}}      & \multicolumn{1}{c|}{\multirow{-3}{*}{{\color[HTML]{333333} PrivFusion(ours)}}} & \multicolumn{1}{c|}{{\color[HTML]{333333} {[}0.01, 0.1, 0.01{]}}}                           & {\color[HTML]{333333} 59.59\%}          & {\color[HTML]{333333} 0.5887}          & {\color[HTML]{333333} 0.5796}          & {\color[HTML]{333333} 63.61\%}          & {\color[HTML]{333333} 59.59\%}          & {\color[HTML]{333333} 66.27\%}          & {\color[HTML]{333333} 68.87\%}          \\ \midrule
\multicolumn{1}{c|}{{\color[HTML]{333333} }}                             & \multicolumn{1}{c|}{{\color[HTML]{333333} Model A}}                       & \multicolumn{1}{c|}{{\color[HTML]{333333} -}}                                               & {\color[HTML]{333333} 95.50\%}          & {\color[HTML]{333333} 0.9549}          & {\color[HTML]{333333} 0.955}           & {\color[HTML]{333333} 95.47\%}          & {\color[HTML]{333333} 95.50\%}          & {\color[HTML]{333333} 95.53\%}          & {\color[HTML]{333333} 95.51\%}          \\
\multicolumn{1}{c|}{{\color[HTML]{333333} }}                             & \multicolumn{1}{c|}{{\color[HTML]{333333} Model B}}                       & \multicolumn{1}{c|}{{\color[HTML]{333333} -}}                                               & {\color[HTML]{333333} 87.94\%}          & {\color[HTML]{333333} 0.8788}          & {\color[HTML]{333333} 0.8797}          & {\color[HTML]{333333} 88.21\%}          & {\color[HTML]{333333} 87.94\%}          & {\color[HTML]{333333} 87.80\%}          & {\color[HTML]{333333} 88.26\%}          \\
\multicolumn{1}{c|}{{\color[HTML]{333333} }}                             & \multicolumn{1}{c|}{{\color[HTML]{333333} Fused Model}}                   & \multicolumn{1}{c|}{{\color[HTML]{333333} -}}                                               & {\color[HTML]{333333} 53.39\%}          & {\color[HTML]{333333} 0.5294}          & {\color[HTML]{333333} 0.5328}          & {\color[HTML]{333333} 52.95\%}          & {\color[HTML]{333333} 53.39\%}          & {\color[HTML]{333333} 52.98\%}          & {\color[HTML]{333333} 53.23\%}          \\ \cmidrule(l){2-10} 
\multicolumn{1}{c|}{{\color[HTML]{333333} }}                             & \multicolumn{1}{c|}{{\color[HTML]{333333} }}                              & \multicolumn{1}{c|}{{\color[HTML]{333333} {[}0.01, 0.1, 1{]}}}                              & {\color[HTML]{333333} \textbf{53.48\%}} & {\color[HTML]{333333} \textbf{0.5227}} & {\color[HTML]{333333} \textbf{0.5284}} & {\color[HTML]{333333} \textbf{52.58\%}} & {\color[HTML]{333333} \textbf{53.48\%}} & {\color[HTML]{333333} \textbf{52.73\%}} & {\color[HTML]{333333} \textbf{52.96\%}} \\
\multicolumn{1}{c|}{{\color[HTML]{333333} }}                             & \multicolumn{1}{c|}{{\color[HTML]{333333} }}                              & \multicolumn{1}{c|}{{\color[HTML]{333333} {[}0.01, 0.1, 0.1{]}}}                            & {\color[HTML]{333333} 51.49\%}          & {\color[HTML]{333333} 0.5134}          & {\color[HTML]{333333} 0.5154}          & {\color[HTML]{333333} 51.36\%}          & {\color[HTML]{333333} 51.40\%}          & {\color[HTML]{333333} 51.35\%}          & {\color[HTML]{333333} 51.62\%}          \\
\multicolumn{1}{c|}{\multirow{-6}{*}{{\color[HTML]{333333} non-IID}}}    & \multicolumn{1}{c|}{\multirow{-3}{*}{{\color[HTML]{333333} PrivFusion(ours)}}} & \multicolumn{1}{c|}{{\color[HTML]{333333} {[}0.01, 0.1, 0.01{]}}}                           & {\color[HTML]{333333} 46.27\%}          & {\color[HTML]{333333} 0.3164}          & {\color[HTML]{333333} 0.2928}          & {\color[HTML]{333333} 50.00\%}          & {\color[HTML]{333333} 46.27\%}          & {\color[HTML]{333333} 23.14\%}          & {\color[HTML]{333333} 21.41\%}          \\ \bottomrule
\end{tabular}
\end{table*}

\subsection{Utility and Privacy Analysis}
\textbf{\textit{Utility Analysis}}. The utility is primarily demonstrated through the acceptance of model fusion results and the reduced reliance on extensive participation of raw data in the PrivFusion framework. However, it is important to consider the impact of the graph matching process on space requirements and overall usability. In the model fusion process, graph matching incurs significant spatial overhead due to the construction of the graph structure and the calculations involved in affinity calculation. This overhead can potentially affect the usability of the PrivFusion framework. However, extensive experimentation and evaluation have shown that PrivFusion achieves satisfactory model fusion results across various settings. Despite the space overhead, PrivFusion manages to strike a balance between privacy preservation and model utility. It successfully preserves privacy without compromising the overall effectiveness of model fusion. Moreover, PrivFusion reduces the reliance on large amounts of original data for model fusion. This is particularly beneficial in scenarios where access to extensive raw data is limited or impractical. By leveraging the shared knowledge and representations in the pre-trained models, the PrivFusion effectively combines models without requiring excessive participation of raw data, thus enhancing usability and reducing data dependency.

In terms of resource consumption, the non-private model fusion has an average processing time of 0.9s, while PrivFusion exhibits a slightly higher average processing time of 1.3s. PrivFusion requires more storage space due to its graph matching process. However, it offers a trade-off by sacrificing some space to achieve faster model updates. This makes PrivFusion suitable for applications that require frequent model updates and prioritize responsiveness over storage efficiency. The increased time overhead in PrivFusion mainly arises from computing activation values during inference and incorporating noise at different stages. Compared to the number of communication rounds typically involved in federated learning, our approach substantially reduces the overall communication overhead.

\textbf{\textit{Privacy Analysis}}. The privacy analysis for PrivFusion focuses primarily on the treatment of model parameters, which undergo standardized processing. Unlike raw data, model parameters have lower sensitivity, and the level of noise introduced during privacy-preserving mechanisms can be controlled. To ensure the privacy of individual parties' models, PrivFusion employs local differential privacy. By incorporating randomized mechanisms like the hybrid-differentially private approach, PrivFusion introduces decentralized federated graph matching with the privacy-preserving affinity calculation process. This noise injection effectively safeguards the privacy of individual model parameters while enabling efficient model fusion. To preserve privacy, PrivFusion avoids direct sharing of raw data and instead capitalizes on the shared knowledge and representations within pre-trained models. This collaborative approach allows parties to benefit from model fusion without disclosing their proprietary data or compromising their privacy. Moreover, the standardized processing of model parameters guarantees that sensitive information remains protected throughout the fusion process.

In contrast, previous researches such as \cite{MSepsilon, Appleepsilon} typically evaluate their systems by testing different values of $\epsilon$ and selecting the minimum value that produces satisfactory results for deployment. For example, Microsoft uses $\epsilon$ = 1 for collecting telemetry data from Windows users \cite{MSepsilon}, while Apple's choice of $\epsilon$ in iOS and macOS ranges from 2 to 8 \cite{Appleepsilon}. However, in our experiments, we have observed that our model remains remarkably robust even when subjected to small $\epsilon$ values below 1. This resilience suggests that our model consistently performs well under stringent privacy constraints, reinforcing its effectiveness in privacy-preserving scenarios.

\subsection{Real-World Applications}
To further investigate the effectiveness of our proposed method PrivFusion in real-world applications, we collected 10,217 color fundus images for diagnosing age-related macular degeneration (shown in Fig. 4) from 17 specialized ophthalmology hospitals in 15 regions. The dataset was split into training, validation, and test sets with an 0.8:0.05:0.15 ratio, respectively. We trained two models with 10 and 50 epochs to fuse in order to simulate the underfitting scenario (sd2t2). To investigate the influence of class imbalance on model fusion performance, 80\% of normal data was randomly removed from the dataset, and one model was trained by removing 70\% of patient data from the same dataset. These two trained models were then fused (unbalanced). In federated learning, different participating parties' data often results in non-IID situations, so we categorized the data by region into two parties, each with data from 7 and 8 regions, and visualized the data in the form of Fig 4. The figures demonstrated significant statistical differences between the datasets from the two region collections. To assess the effectiveness of our method under non-IID conditions, we trained a model separately with data from each participating party and conducted a model fusion experiment (non-IID).

\begin{figure}[t]
\includegraphics[width=\linewidth]{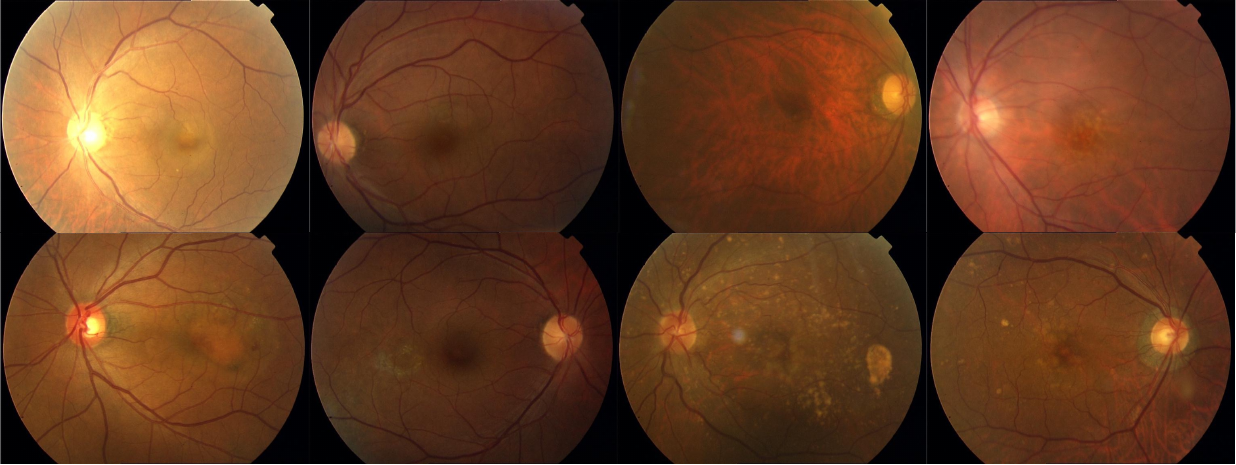}
\caption{Typical color fundus images in our age-related macular degeneration dataset.}
\label{figeye}
\end{figure}

\begin{figure}[ht]
\begin{minipage}[b]{0.45\linewidth}
\centering
\includegraphics[width=\textwidth]{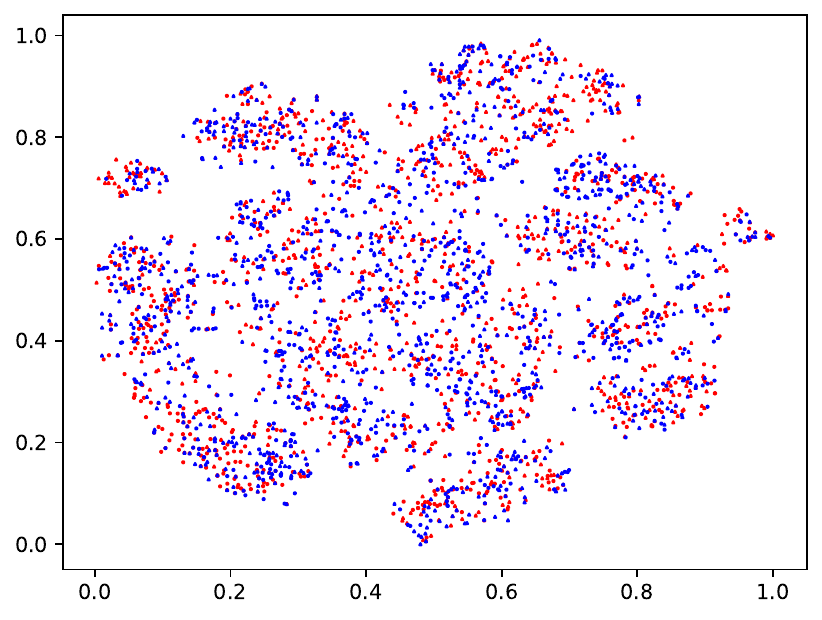}
\label{fig:figure4}
\end{minipage}
\hspace{0.5cm}
\begin{minipage}[b]{0.45\linewidth}
\centering
\includegraphics[width=\textwidth]{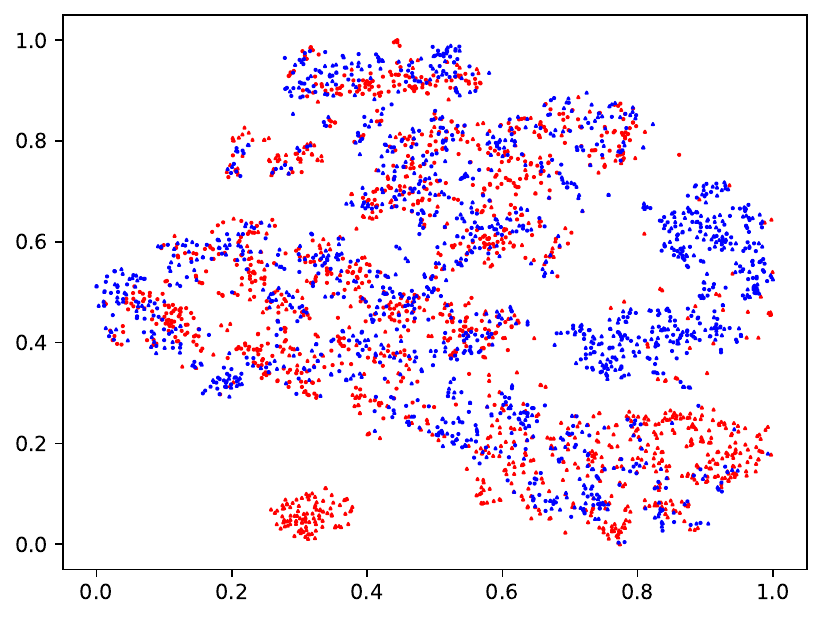}
\label{fig:figure2}
\end{minipage}
\caption{The homogeneous and heterogeneous distribution of real application dataset.}
\end{figure}

Fig. 5 presents the true distribution of the training dataset, where the left plot represents an IID scenario, while the right plot portrays a non-IID scenario. The red and blue colors indicate distinct datasets, with circles and triangles representing different classes. In the left plot, both datasets exhibit a high degree of similarity, indicating a significant overlap in their distributions. Consequently, when integrating or combining models, we can expect improved performance since the datasets possess comparable characteristics. Conversely, the right plot demonstrates substantial dissimilarity and dispersion in the distributions of the two datasets. This discrepancy suggests the presence of diverse patterns and distinct variations between the datasets. Integrating models trained on such disparate datasets presents a greater challenge. The disparities in distribution can introduce inconsistencies in learned representations and decision boundaries, thereby impeding effective model fusion and optimal performance. Hence, the analysis of Fig. 5 underscores the significance of considering the distribution of training data when performing model fusion or combination. When datasets share similar distributions, fusion processes are likely to yield superior outcomes. Conversely, when significant distributional differences exist, fusion tasks become more intricate, necessitating meticulous consideration and potentially specialized techniques to address the disparities.

In Table IV, we present the experimental results under non-homogeneous, underfitting, and non-IID settings. We evaluated the model fusion using metrics such as accuracy (Acc), macro F1-score (ma\_F1), weighted F1-score (w\_F1), macro recall (ma\_Rec), weighted recall (w\_Rec), macro precision (ma\_Prec), and weighted precision (w\_Prec). The results demonstrate that PrivFusion achieves satisfactory performance across all these metrics. PrivFusion effectively addresses the challenges associated with non-homogeneous data, underfitting, and non-IID distribution. It successfully combines models from multiple parties while preserving privacy and maintaining good utility. The measured metrics consistently indicate high performance in terms of accuracy, F1 score, recall, and precision.

Moreover, PrivFusion allows users to adjust the privacy budget, enabling a trade-off between utility and privacy. By varying the privacy budget, users can fine-tune the level of privacy protection while ensuring acceptable utility. The results in Table IV confirm the effectiveness of PrivFusion in model fusion under diverse and challenging data settings. It achieves a balance between utility and privacy, facilitating secure collaboration and model fusion among multiple parties with satisfactory performance across various evaluation metrics. These findings highlight the practicality and effectiveness of PrivFusion in real-world scenarios.

\section{Conclusion}
Decentralized model fusion has emerged as a promising approach for aggregating models while preserving privacy in collaborative settings. In this paper, we present PrivFusion, a novel architecture for privacy-preserving decentralized model fusion. PrivFusion addresses the challenges of sharing trained models under local differential privacy constraints, ensuring data confidentiality and secure collaboration. Our method leverages a graph-based representation of the neural network and employs randomized mechanisms for decentralized federated graph matching to protect the privacy of neuron alignment during model fusion. We analyze the impact of different privacy budgets on model fusion and highlight the significant contribution of activation values to the fused model's performance. Moreover, our findings demonstrate the advantages of adopting a hybrid LDP approach, which combines the benefits of LDP mechanisms for preserving privacy during the fusion process. Experimental results on two image datasets and real-world healthcare applications validate the effectiveness of PrivFusion in preserving privacy while maintaining model utility.

In conclusion, PrivFusion offers a valuable solution for privacy-preserving decentralized model fusion, with potential applications in various domains. Future research can explore the scalability of PrivFusion and extend its use to large-scale pre-trained models in collaborative scenarios. Additionally, investigating noise-adaptive perturbation techniques is another avenue for future research. The advancements made in this paper contribute to the field of privacy-preserving decentralized model fusion, enabling secure and collaborative data analysis.




 
%

\bibliographystyle{IEEEtran}
\bibliography{ref}  
\vfill

\end{document}